\documentclass[12pt]{article}

\usepackage{lmodern}            
\usepackage{amsmath,amssymb}    
\usepackage{amsthm}             
\usepackage{authblk}            
\usepackage{geometry}
\usepackage{graphicx}
\usepackage{siunitx}
\usepackage{tabularx,booktabs,makecell}
\usepackage{ltablex} \keepXColumns
\usepackage{algorithm}
\usepackage{algpseudocode}
\usepackage{enumitem}
\usepackage{caption}
\usepackage[authoryear,round]{natbib}

\usepackage[title]{appendix}
\usepackage{hyperref}           

\newtheorem{theorem}{Theorem}[section]
\newtheorem{proposition}[theorem]{Proposition}
\newtheorem{lemma}[theorem]{Lemma}
\newtheorem{corollary}[theorem]{Corollary}
\theoremstyle{remark}
\newtheorem{remark}[theorem]{Remark}
\theoremstyle{definition}
\newtheorem{definition}[theorem]{Definition}
\newtheorem{example}[theorem]{Example}

\title{Sign-Aware Multistate Jaccard Kernels and Geometry for Real and Complex-Valued Signals}
\author[1]{Vineet Yadav}
\affil[1]{Jet Propulsion Laboratory, California Institute of Technology, Pasadena, CA, USA}
\date{} 

\begin{document}
\maketitle

\begin{abstract}
We introduce a sign-aware, multistate Jaccard/Tanimoto framework that extends overlap-based distances from nonnegative vectors and measures to arbitrary real- and complex-valued signals while retaining bounded metric and positive-semidefinite kernel structure. Formally, the construction is a set- and measure-theoretic geometry: signals are represented as atomic measures on a signed state space, and similarity is given by a generalized Jaccard overlap of these measures. Each signal is embedded into a nonnegative multistate representation, using positive/negative splits for real signals, Cartesian and polar decompositions for complex signals, and user-defined state partitions for refined regime analysis. Applying the Tanimoto construction to these embeddings yields a family of $[0,1]$ distances that satisfy the triangle inequality and define positive-semidefinite kernels usable directly in kernel methods and graph-based learning. Beyond pairwise distances, we develop coalition analysis via M\"obius inversion, which decomposes signal magnitude into nonnegative, additive contributions with exact budget closure across coalitions of signals. Normalizing the same embeddings produces probability measures on coordinate--state configurations, so that the distance becomes a monotone transform of total variation and admits a regime--intensity decomposition. The resulting construction yields a single, mechanistically interpretable distance that simultaneously provides bounded metric structure, positive-semidefinite kernels, probabilistic semantics, and transparent budget accounting within one sign-aware framework, supporting correlograms, feature engineering, similarity graphs, and other analytical tools in scientific and financial applications.

\end{abstract}

\section{Introduction}
\label{sec:introduction}

Similarity measures and kernels constitute the geometry of modern machine
learning, governing the performance of nearest-neighbor methods, kernel
machines, Gaussian processes, and representation learning. In these
settings, a rigorous notion of ``closeness'' must satisfy dual constraints:
it must be mathematically well behaved (e.g., satisfy metric axioms or
yield positive-semidefinite kernels) while remaining scientifically meaningful
for the domain at hand.

Two distinct families of comparison dominate current practice. The first
comprises geometric distances such as Euclidean and Mahalanobis distance~\citep{mahalanobis1936generalised},
cosine similarity~\citep{salton1983introduction}, and dynamic time warping~\citep{sakoe1978dynamic}. These methods treat data as
vectors in a continuous linear space, quantifying proximity through
coordinate-wise error aggregation or angular alignment. They perform well
when a signal is defined by its location and direction in $\mathbb{R}^n$,
but they lack an intrinsic notion of set-theoretic overlap or mass
conservation.

The second family descends from the Jaccard index~\citep{Jaccard1901} and a
sequence of related overlap coefficients that were originally formulated on
incidence or count data, not on an explicit notion of ``mass''. Early
set-based variants such as the coefficients of
Kulczy\'{n}ski and the binary forms used by Rogers and Tanimoto
\citep{kulczynski1927pflanzenassoziationen,rogers_tanimoto_1960,tanimoto1958elementary}
operate on incidence vectors in $\{0,1\}^n \subset \mathbb{Z}^n$.
Closely related overlap functionals for abundance or count tables were
introduced by Czekanowski and Motyka in $\mathbb{N}^n$
\citep{czekanowski1909zur,motyka1947o}, and Ru\v{z}i\v{c}ka made the extension
to general nonnegative vectors $x \in \mathbb{R}_{\ge 0}^n$ explicit
\citep{Ruzicka1958}. Later, in computer vision, the same algebraic
intersection pattern was applied to color histograms in
$\mathbb{R}_{\ge 0}^n$ by Swain and Ballard~\citep{swain1991color}, again as a
similarity between empirical histograms rather than as a measure-theoretic
construction.

In all of these classical formulations, the index is a two-argument
functional on a nonnegative sequence space,
\[
  S : \mathcal{X}^n \times \mathcal{X}^n \to [0,1],
\]
where $\mathcal{X}$ is typically $\{0,1\}$, $\mathbb{N}$, or
$\mathbb{R}_{\ge 0}$ depending on the application. Analyses involving many
samples proceed via pairwise similarity matrices built from $S(A,B)$, rather
than through a single multi-way or coalition-level overlap functional.

While pairwise similarities $S(A,B)$ are sufficient for many tasks, a
growing class of problems demands \emph{multi-way} comparison and
budgeting. Examples include comparing ensembles of models for the same
physical field, decomposing profit and loss across multiple firms or
portfolios, or attributing precipitation across overlapping events or
regions. In such settings, it is not enough to know that signals are
pairwise similar; one needs a coalition-level decomposition that explains
how total magnitude is shared among groups of signals.

In this paper we \emph{reinterpret} the overlap formulas on
$\mathbb{R}_{\ge 0}^n$ as acting on discrete mass distributions supported on a
finite partition of a subset of $\mathbb{C}$, with masses taking values in
$\mathbb{R}_{\ge 0}$. This mass-based, set- and measure-theoretic viewpoint is
not present in the original references; it is the geometric formalization of
these indices that we develop in this paper.

Because they retain the physical units of the data, such overlap measures are
ubiquitous in ecology~\citep{bray1957ordination},
cheminformatics~\citep{bajusz2015tanimoto}, and other fields like textual similarity~\citep{broder1997resemblance} where absolute
quantities carry meaning. However, existing
analyses of the Jaccard and Tanimoto indices are predominantly restricted to
sets, measures, and nonnegative vectors representing counts, intensities, or
membership weights; they neither accommodate signed or complex signals nor
provide explicit probabilistic or coalition-level budget interpretations.
Historically, this limitation reflects the fact that these indices were
designed to quantify overlap in the \emph{presence} and \emph{magnitude} of a
phenomenon (e.g., species abundances, molecular fragments, shingle
frequencies), not gains versus losses. Consequently, there was no notion of
``mass loss'' or negative contribution at a coordinate, so negative or
complex-valued entries simply did not arise in the modeling space.

Real-world signals, however, are rarely nonnegative. Positive and negative excursions typically encode different mechanisms: above- versus below-baseline flux anomalies, model under- versus over-prediction in residual analysis, upward versus downward moves in asset prices, gene up- versus down-regulation, or the real and imaginary parts of a complex waveform. Existing overlap measures, designed for nonnegative inputs, cannot distinguish whether two vectors swing in the same or opposite directions: they collapse $+x$ and $-x$ to the same absolute value. Correlation and cosine similarity, in turn, identify antipodal signals as ``maximally similar,'' even though one is just a sign flip of the other. None of these indices directly answers the question:

\begin{center}
\emph{How much do two signed (or complex) signals agree once we pay attention to direction and regime?}
\end{center}

From a modeling perspective this question arises immediately. In virtually any learning or inference pipeline we can write
\begin{equation}
z = M_\theta(x) + e,
\end{equation}
where $z$ denotes data, $M_\theta$ is a prediction or generative mechanism, and $e$ is a residual or discrepancy signal. It is the structure of $e$ that reveals how the model succeeds or fails: sustained positive excursions indicate systematic underprediction, negative excursions indicate overprediction, and deviations concentrated in particular regimes or time windows signal structural change or missing mechanisms. A single scalar loss $\|e\|$ or a correlation coefficient collapses this structure.

\paragraph{Desiderata.}
The aim of this paper is not to add one more distance to an already long
list~\citep[see, e.g.,][Tables~1--9]{cha2007comprehensive}, but to obtain a single, reusable construction that simultaneously:
\begin{enumerate}[label=(D\arabic*)]
    \item is sign- and regime-aware: it distinguishes agreement in direction and between scientifically meaningful states (e.g., large loss / small loss / neutral / small gain / large gain);
    \item yields a bounded $[0,1]$ metric and a positive-semidefinite kernel that plugs directly into standard ML machinery;
    \item supports exact budget accounting: nonnegative, additive decompositions of shared magnitude (or energy/occupancy) across coordinates and coalitions of signals, with no residual;
    \item admits a clear probabilistic semantics, so that distances can be interpreted as total-variation discrepancies between induced measures;
    \item extends naturally to complex-valued embeddings without losing any of the above.
\end{enumerate}

Standard families offer some of these properties in isolation. For example, Wasserstein distances provide probabilistic interpretation~\citep{villani2009optimal}, Jaccard/Tanimoto indices provide unit-preserving overlap~\citep{bray1957ordination}, and many kernels are PSD by design~\citep{Scholkopf2002}. To our knowledge, however, none provides all of (D1)--(D5) within a single model-agnostic framework tailored to signed and complex signals.

\paragraph{Contributions.}
This paper makes two main contributions.

\begin{enumerate}[label=(\roman*)]
    \item We lift the classical Jaccard/Tanimoto overlap family from nonnegative vectors and measures to arbitrary real- and complex-valued signals via sign-aware embeddings, while preserving bounded metric structure and positive-semidefinite kernel representations.
    
    \item We equip this overlap family with new probabilistic and budget interpretations: every embedded signal induces a normalized measure on coordinate--state atoms, so the resulting distance is a monotone transform of total variation, and M\"obius inversion yields an exact nonnegative decomposition of $L^1$ magnitude across coalitions of signals.
\end{enumerate}

We introduce a sign-aware, multistate Jaccard--Tanimoto framework that extends overlap-based thinking to real and complex signals whose sign and regime matter. The construction is based on a multistate embedding $\psi^{(K)}(A)$, which maps a signal $A$ into a nonnegative tensor whose entries record a nonnegative moment (e.g., magnitude, energy) of $A$ lying in each state at each index. States are user-defined intervals $B_0, \ldots, B_{K-1}$ on the real line (e.g., a neutral ``noise'' band plus moderate and extreme positive/negative deviations), and for complex signals we provide both Cartesian and polar decompositions. Applying the classical Tanimoto construction on these nonnegative embeddings yields a family of $[0,1]$ distances $d_{\text{peak}}$ that satisfy the triangle inequality and define positive-semidefinite kernels usable directly in kernel methods and graph-based learning. Normalizing the same embeddings produces probability measures on
coordinate--state configurations, and inclusion--exclusion (M\"obius
inversion; see \citet{rota1964mobius}) then decomposes magnitude into
nonnegative, additive coalition budgets with exact closure.

Taken together, these properties provide a unified infrastructure for bounded metrics, positive-semidefinite kernels, probabilistic semantics, transparent budget accounting, and built-in regime and intensity separation, supporting sign-aware correlograms, feature engineering, similarity graphs, and coalition diagnostics in scientific, financial, and other machine learning applications.

The remainder of the paper proceeds as follows. Section~\ref{sec:overview_metrics} reviews existing metrics and kernels and introduces notation. Section~\ref{sec:core_math_properties} formalizes the sign-aware and multistate constructions, proves that the resulting distances are metrics and that the associated similarities define positive-semidefinite kernels, and extends the framework to complex-valued signals. Section~\ref{sec:untrimmed_multistate} develops the coalition (Venn/M\"obius) budgeting and probabilistic views. Section~\ref{sec:visualization_case_study} illustrates the behavior of the distance and kernel on synthetic examples and sketches how the framework can serve as a base layer for sign-aware correlograms, variograms, feature engineering, and similarity graphs. Section~\ref{sec:why_distinctive} compares the peak-to-peak family with established distance measures and summarizes the distinctive combination of properties achieved by the proposed framework.

\section{Overview of Metrics and Concepts}\label{sec:overview_metrics}

Table~\ref{tab:key_symbols} provides a compact reference for notation used through Section~\ref{sec:untrimmed_multistate}.
Unnormalized quantities ($N$, $U_{\text{peak}}$) preserve the physical units of the data, while normalized
quantities ($J_{\text{peak}}$, $d_{\text{peak}}$) are dimensionless.

\begin{longtable}{@{}>{\raggedright\arraybackslash}p{2.8cm}p{7.8cm}>{\raggedright\arraybackslash}p{3.0cm}@{}}
\caption{Key notation and mathematical objects (through Section~\ref{sec:untrimmed_multistate}).}
\label{tab:key_symbols}\\
\toprule
\textbf{Symbol} & \textbf{Description} & \textbf{Reference} \\
\midrule
\endfirsthead

\multicolumn{3}{l}{\small\itshape Table \thetable\ (continued)}\\
\toprule
\textbf{Symbol} & \textbf{Description} & \textbf{Reference} \\
\midrule
\endhead

\midrule
\multicolumn{3}{r}{\small\itshape Continued on next page}\\
\endfoot

\bottomrule
\endlastfoot

\multicolumn{3}{l}{\textbf{Sign structure}}\\[3pt]
$\operatorname{sgn}(x)$ &
Sign function taking values in $\{-1,0,+1\}$ &
Def.~\ref{def:epsilon0}\\[6pt]

\multicolumn{3}{l}{\textbf{Sign-split embedding}}\\[3pt]
$x_i^+,\,x_i^-$ &
Positive and negative parts:
$x_i^+=\max\{x_i,0\}$, $x_i^-=\max\{-x_i,0\}$ &
Def.~\ref{def:phi_embedding}\\
$\phi(X)$ &
Sign-split embedding $\phi:\mathbb{R}^n\to\mathbb{R}^{2n}_{\ge 0}$ &
Def.~\ref{def:phi_embedding}\\[6pt]

\multicolumn{3}{l}{\textbf{Peak intersection, union, and similarity}}\\[3pt]
$\mathcal{I}_{\text{same}}(A,B)$ &
Index set where $A$ and $B$ share the same nonzero sign &
Def.~\ref{def:index_same_sign}\\
$N(A,B)$ &
Sign-aware intersection (units preserved): overlap of same-signed excursions &
Def.~\ref{def:pairwise_intersection}\\
$U_{\text{peak}}(A,B)$ &
Peak-to-peak union (units preserved): sum of half-wave maxima for the pair &
Def.~\ref{def:peak_union}\\
$J_{\text{peak}}(A,B)$ &
Peak Jaccard coefficient:
$N(A,B)/U_{\text{peak}}(A,B)\in[0,1]$ (defined as $1$ when $U_{\text{peak}}{=}0$) &
Def.~\ref{def:peak_similarity}\\
$d_{\text{peak}}(A,B)$ &
Peak distance: $1- J_{\text{peak}}(A,B)\in[0,1]$ &
Def.~\ref{def:peak_distance}\\
$K_{\text{peak}}(A,B)$ &
Peak kernel: $K_{\text{peak}}{=}J_{\text{peak}}$ is positive-semidefinite &
Thm.~\ref{thm:pd_peak_kernel}\\[6pt]

\multicolumn{3}{l}{\textbf{Multistate embedding and coalitions}}\\[3pt]
$\mathcal{B}$ &
Partition of $\mathbb{R}$: $\{B_0,\ldots,B_{K-1}\}$ used in multistate embeddings &
Def.~\ref{def:multistate_partition}\\
$\psi^{(K)}(X)$ &
Multistate norm-preserving embedding
$\psi^{(K)}:\mathbb{R}^n\to\mathbb{R}_{\ge 0}^{n\times K}$ with respect to $\mathcal{B}$ &
Def.~\ref{def:multistate_partition}\\
$N(S)$ &
Cumulative multistate intersection (coalition budget) for $S\subseteq[m]$ &
Def.~\ref{def:cumulative_untrimmed}\\
$\widetilde{N}(S)$ &
Exclusive multistate intersection (Möbius-inverted budgets) for coalition $S$ &
Def.~\ref{def:exclusive_untrimmed}\\
$\Omega$ &
Coordinate--state space indexing multistate atoms &
Def.~\ref{def:coordinate_state_space}\\
$\nu_A$ &
Finite measure on $\Omega$ induced by signal $A$ &
Def.~\ref{def:magnitude_measure}\\
$P_A$ &
Normalized probability distribution on $\Omega$ induced by $A$ &
Def.~\ref{def:normalized_probability}\\[6pt]

\multicolumn{3}{l}{\textbf{Metric and kernel properties}}\\[3pt]
& Metric property: $d_{\text{peak}}{=}d_{\mathrm{Tan}}(\phi(A),\phi(B))$ &
Thm.~\ref{thm:metric_dpeak}\\
& Positive-semidefiniteness of $K_{\text{peak}}$ &
Thm.~\ref{thm:pd_peak_kernel}\\
$d_{\mathrm{multi}}(X,Y)$ &
Multistate Tanimoto distance
$d_{\mathrm{Tan}}(\psi^{(K)}(X),\psi^{(K)}(Y))$; metric on $\psi^{(K)}(\mathbb{R}^n)$, pseudometric on $\mathbb{R}^n$ &
Thm.~\ref{thm:multistate_metric}\\
$K_{\mathrm{multi}}(X,Y)$ &
Multistate Tanimoto kernel
$J_{\mathrm{Tan}}(\psi^{(K)}(X),\psi^{(K)}(Y))$, positive-semidefinite on $\mathbb{R}^n$ &
Thm.~\ref{thm:multistate_metric}\\[6pt]

\multicolumn{3}{l}{\textbf{Generalized embeddings}}\\[3pt]
$\psi_{\mathrm{C}}(Z)$ &
Cartesian embedding for complex signals (real/imaginary sign-split) &
Def.~\ref{def:complex_embeddings}\\
$\psi_{\mathrm{P}}^{(K)}(Z)$ &
Polar embedding for complex signals (magnitude with phase partition) &
Def.~\ref{def:complex_embeddings}\\[6pt]

\multicolumn{3}{l}{\textbf{Classical references}}\\[3pt]
$d_{\mathrm{Tan}},\,J_{\mathrm{Tan}}$ &
Tanimoto distance and similarity on nonnegative vectors;
$d_{\mathrm{Tan}}(x,y)=1-J_{\mathrm{Tan}}(x,y)$ and
$d_{\text{peak}}{=}d_{\mathrm{Tan}}\!\circ\!\phi$ on embedded signals &
Thm.~\ref{thm:metric_dpeak}, Lem.~\ref{lem:Jpeak_is_Tanimoto}\\

\end{longtable}

\paragraph{Computational verification.}
To complement the analytic development in Sections~\ref{sec:core_math_properties}
and~\ref{sec:untrimmed_multistate}, we implemented the full peak-to-peak
framework in Python, including the sign-split embedding $\phi$, the multistate
embedding $\psi^{(K)}$, coalition Möbius inversion, the probabilistic
TV/$\delta$ representation, and the complex Cartesian extension $\psi_C$.
For every worked example in the manuscript (Eqs.~(39)–(40),(47)–(48),(57)–(65),(70)–(72), and(113)–(128)), the implementation reproduces the reported numerical values exactly.
Monte Carlo experiments further confirm that
(i) $d_{\text{peak}}(A,B)$ coincides with the min--max Tanimoto distance on
$\phi(A),\phi(B)$;
(ii) the real and complex kernels
$K(A,B) = J_{\text{peak}}(A,B)$ and
$K_{\psi_C}(Z_1,Z_2)$ produce positive-semidefinite Gram matrices;
(iii) $d_{\text{peak}}$ satisfies the triangle inequality and is invariant
under common positive rescaling; and
(iv) coalition budgets $\sum_{S \ni j} \tilde N(S)$ close numerically to
$\|A_j\|_1$, and coarsening the state partition decreases the distance,
in line with the data-processing interpretation.
While not a substitute for formal proofs, these checks provide an independent,
executable confirmation that the algebraic identities and structural
properties in the sequel are correctly implemented and numerically stable.

\section{Core Mathematical Properties}
\label{sec:core_math_properties}

This section establishes the mathematical foundations of the sign-aware similarity framework, proving that the proposed measures constitute a proper metric and demonstrating their extension to multi-way comparisons.

\subsection{Design axioms for signed excursions}
\label{sec:design_axioms}

Our construction ultimately applies to complex-valued signals
$X \in \mathbb{C}^n$, but it operates on the real-valued excursions that
arise from their real and imaginary parts. This decomposition is not merely
algebraic; it respects the phenomenological reality that orthogonal components
(real vs.\ imaginary) and opposing polarities (positive vs.\ negative) often
represent distinct physical channels—such as in-phase vs.\ quadrature states or
assets vs.\ liabilities—where structural alignment is semantically distinct
from arithmetic cancellation. In this subsection we therefore formulate the
basic requirements that any sign-aware overlap and union should satisfy for
real-valued channels, and we state the axioms for $A,B \in \mathbb{R}^n$. The
extension to complex-valued signals $X \in \mathbb{C}^n$ is obtained by
applying the same construction separately to $\Re X$ and $\Im X$; coalition
analyses for $(A_1, \ldots, A_m)$ are then performed by treating these real and
imaginary components as distinct channels within the unified embedding space.

\begin{enumerate}[label=(A\arabic*)]
  \item \textbf{Sign locality.}
  Overlap and union are computed coordinatewise: the contribution at
  coordinate $i$ depends only on the pair $(A_i,B_i)$ and is independent of
  all other coordinates.

  \item \textbf{Excursion separation.}
  Positive and negative excursions are treated as disjoint channels.
  A coordinate with opposite signs, $A_i B_i < 0$, contributes zero overlap:
  only excursions with the same nonzero polarity can reinforce each other.

  \item \textbf{Magnitude monotonicity.}
  The overlap and union functionals should be non-decreasing with respect to
  magnitude at each coordinate. Specifically, if $|A_i|$ increases while its
  sign and $B_i$ are held fixed, then the local contributions to the overlap
  $N(A,B)$ and the union $U_{\text{peak}}(A,B)$ at coordinate $i$ should not
  decrease.

  \item \textbf{$L^1$ compatibility and reduction.}
  The total ``mass'' of a signal is measured by its $L^1$ norm,
  $\|A\|_1 = \sum_i |A_i|$. The overlap and union must respect this scale,
  satisfying $0 \le N(A,B) \le U_{\text{peak}}(A,B) \le \|A\|_1 + \|B\|_1$.
  Furthermore, when $A,B \in \mathbb{R}_{\ge 0}^n$, the construction must
  reduce to the classical intersection and union operators underlying the
  Tanimoto and Jaccard indices.
\end{enumerate}

The sign-split embedding in Definition~\ref{def:phi_embedding}, together
with the overlap $N(A,B)$ in~\eqref{eq:N_AB_def} and the peak-to-peak union
$U_{\text{peak}}(A,B)$ in~\eqref{eq:U_peak_def}, constitutes a natural and
minimal construction that satisfies axioms~(A1)--(A4) while preserving the
physical units of the data.

\begin{remark}[Zero as a physical state]
\label{rem:zero_state}
Our treatment of zero-valued excursions follows the same physical design
principles. Zero is not interpreted as an algebraic indeterminacy of the
form ``$0/0$'' arising in a normalized index; it is a bona fide physical
state. Two identically zero signals correspond to the same state of the
system and are therefore regarded as perfectly similar. Consistent with the
normalization of the overlap and union, the resulting peak-to-peak
similarity satisfies
\[
  J_{\text{peak}}(0,0) = 1.
\]
This behavior is physically meaningful in applications where zero encodes a
distinct regime, for example:
\begin{itemize}
  \item Mass $=0$: ``no mass present'' in a control volume;
  \item Flux $=0$: ``perfect balance'' with no net inflow or outflow;
  \item Rain $=0$: ``drought conditions'' with complete absence of
        precipitation.
\end{itemize}
In all such cases, the absence of activity is a well-defined state in the
geometry rather than an undefined limit.
\end{remark}

\begin{remark}[Semantic separation of channels]
\label{rem:semantic_channels}
Axiom~(A2) encodes the intuition that overlap is only meaningful within the
same semantic channel. Positive excursions typically represent assets,
gains, or production (e.g., portfolio value, gross primary productivity,
revenue), whereas negative excursions represent losses, costs, or
consumption (e.g., drawdowns, respiration, expenses). In such settings,
it is natural to assert that ``assets overlap with assets'' and ``losses
overlap with losses,'' but not that an asset overlaps with a loss. The
sign-split embedding enforces this separation: only like-signed
contributions reinforce each other in the overlap $N(A,B)$, while
opposite-signed excursions are accounted for via the union
$U_{\text{peak}}(A,B)$ and the resulting distance.
\end{remark}

\subsection{Foundational Definitions}
\label{sec:foundational_defs}

We begin by formalizing the embedding that enables sign-aware comparisons while maintaining mathematical rigor.

\begin{definition}
\label{def:epsilon0}
For any real number $x \in \mathbb{R}$, we define the sign function as
\begin{equation}
  \operatorname{sgn}(x)
  := 
  \begin{cases}
    +1 & \text{if } x > 0, \\[2pt]
     0 & \text{if } x = 0, \\[2pt]
    -1 & \text{if } x < 0.
  \end{cases}
\end{equation}
This function serves as the coordinate-wise mechanism for axiom \textbf{A1}: throughout this work, all comparisons and sign determinations employ this function locally at each index $i$. Consequently, the metric properties established in \autoref{sec:metric_property} emerge from the aggregation of these independent local operations.
\end{definition}

\begin{definition}[Sign-split embedding]
\label{def:phi_embedding}
The sign-split embedding $\phi(X)$ realizes the \textbf{Excursion separation} axiom \textbf{A2} by transforming a real-valued signal into a non-negative representation that strictly separates positive and negative components. For a vector $X=(x_1,\ldots,x_n)\in\mathbb{R}^n$, we define the embedding $\phi: \mathbb{R}^n \to \mathbb{R}_{\ge 0}^{2n}$ as
\begin{equation}
  \phi(X)
  := 
  (x_1^{+},x_1^{-},x_2^{+},x_2^{-},\ldots,x_n^{+},x_n^{-}),
\label{eq:def_phi}
\end{equation}
where for each coordinate $i$,
\begin{equation}
  x_i^{+}:=\max\{x_i,0\},\qquad
  x_i^{-}:=\max\{-x_i,0\}.
\end{equation}
This decomposition ensures that each component $x_i$ is represented by exactly one non-zero value in the pair $(x_i^{+}, x_i^{-})$, enforcing the disjoint channel requirement of axiom \textbf{A2}. When $x_i > 0$, the positive part $x_i^{+}$ captures its magnitude while $x_i^{-} = 0$. Conversely, when $x_i < 0$, the negative part $x_i^{-}$ records $|x_i|$ while $x_i^{+} = 0$. This alternating structure $(+, -)$ creates a doubled space where sign information is encoded through position rather than through the numerical sign itself. For example, if $x_i = -3$, then $(x_i^{+}, x_i^{-}) = (0, 3)$. If $x_i = 5$, then $(x_i^{+}, x_i^{-}) = (5, 0)$.
\end{definition}

A key property of this embedding is that it preserves the signal's total magnitude, which is crucial for maintaining the physical units of the data. The following lemma formally establishes that this construction satisfies the mass conservation axiom.

\begin{lemma}[$L^{1}$-preservation of the sign-split embedding]
\label{lem:l1_preserve}
The sign-split embedding preserves the total magnitude of the signal. For every $X\in\mathbb{R}^n$,
\begin{equation}
  \|\phi(X)\|_{1} = \|X\|_{1}.
\end{equation}
\end{lemma}

\begin{proof}
By construction, the positive and negative parts satisfy the identity $|x_i| = x_i^{+} + x_i^{-}$ for every coordinate $i$. Summing over all coordinates yields:
\begin{equation}
   \|\phi(X)\|_{1}
   = \sum_{i=1}^{n}(x_i^{+}+x_i^{-})
   = \sum_{i=1}^{n}|x_i|
   = \|X\|_{1}.
\end{equation}
This explicitly confirms that the embedding $\phi$ satisfies the \textbf{$L^1$ compatibility} requirement of axiom \textbf{A4}
\end{proof}

\begin{remark}
The preservation of the $L^1$ norm ensures that physical units and scale are maintained throughout the embedding process, as demanded by axiom \textbf{A4}. This property is also essential for the isometric relationship established in Theorem~\ref{thm:metric_dpeak}.
\end{remark}

\subsection{Pairwise Sign-Aware Measures}
\label{sec:pairwise_measures}

We now construct the specific overlap and union functionals that satisfy the design axioms of Section~\ref{sec:design_axioms}. By applying the classical intersection and union logic to the sign-split embedding $\phi(A)$, we obtain three building-block quantities: (i) a \emph{sign-aware intersection} that realizes the excursion separation axiom, (ii) a \emph{peak-to-peak union} that captures the full dynamic range, and (iii) the normalized similarity and distance derived from their ratio.

\subsubsection{Sign-aware intersection}
\label{sec:pairwise_sign_aware}

To satisfy the \textbf{Excursion separation} axiom \textbf{A2}, the intersection measure must quantify shared magnitude only where signals maintain directional consistency. This requires identifying coordinates where reinforcing polarities occur.

\begin{definition}[Index set of sign agreement]
\label{def:index_same_sign}
For two signals $A, B \in \mathbb{R}^{n}$, the index set of sign agreement comprises all coordinates where both signals exhibit the same non-zero polarity:
\begin{equation}
  \mathcal{I}_{\text{same}}(A,B)
  \;=\;
  \bigl\{\,i\in\{1,\dots,n\}\; \bigl| \;
      \operatorname{sgn}(A_i)
      \;=\;
      \operatorname{sgn}(B_i)
      \;\neq\; 0
  \bigr\}.
\label{eq:index_same_sign}
\end{equation}
This set isolates the domain where signal reinforcement is physically meaningful, filtering out opposing contributions.
\end{definition}

\begin{definition}[Pairwise sign-aware intersection]
\label{def:pairwise_intersection}
The sign-aware intersection $N(A,B)$ is defined as the component-wise minimum of the sign-split embeddings, restricted to the set of sign agreement:
\begin{equation}
  N(A,B)
  :=
  \sum_{i\in\mathcal{I}_{\text{same}}(A,B)}
  \min\bigl(|A_i|, |B_i|\bigr).
\label{eq:N_AB_def}
\end{equation}
This formulation enforces axiom \textbf{A2}: if $A_i$ and $B_i$ have opposite signs, $i \notin \mathcal{I}_{\text{same}}$ and the contribution is zero. Consistent with axiom \textbf{A4}, the measure preserves the physical units of the signal, representing the total magnitude shared by $A$ and $B$ within the same semantic channel.
\end{definition}

\subsubsection{Peak-to-peak union}
\label{sec:peak_union}

To normalize the intersection, we require a union measure that respects the \textbf{Magnitude monotonicity} axiom \textbf{A3} bounds the $L^1$ norm. This is achieved by computing the envelope of the excursions in the sign-split space.

\begin{definition}[Peak-to-peak union]
\label{def:peak_union}
For signals $A, B \in \mathbb{R}^n$, the peak-to-peak union is defined as the sum of the coordinate-wise maxima of the positive and negative parts:
\begin{equation}
  U_{\text{peak}}(A,B)
  :=
  \sum_{i=1}^{n}\Bigl(\max\{A_i^{+},B_i^{+}\}
                      + \max\{A_i^{-},B_i^{-}\}\Bigr).
\label{eq:U_peak_def}
\end{equation}
\end{definition}

By taking the maximum of the sign-split components, $U_{\text{peak}}$ captures the full dynamic range spanned by the signal pair. This definition satisfies axiom \textbf{A3} (since the maximum is monotonic) and axiom \textbf{A4} ($L^1$ compatibility) via the inequality $\max(x,y) \le x+y$ for non-negative inputs, ensuring $U_{\text{peak}}(A,B) \le \|A\|_1 + \|B\|_1$.

\subsubsection{Similarity coefficient and distance}
\label{sec:peak_similar}

The ratio of the sign-aware intersection to the peak-to-peak union yields a normalized similarity coefficient analogous to the classical Jaccard index but adapted for signed data.

\begin{definition}[Peak-to-peak similarity coefficient]
\label{def:peak_similarity}
The peak-to-peak similarity coefficient between signals $A$ and $B$ is
\begin{equation}
  J_{\text{peak}}(A,B)
  :=
  \begin{cases}
     \dfrac{N(A,B)}{U_{\text{peak}}(A,B)}, & \text{if } U_{\text{peak}}(A,B) > 0,\\[6pt]
     1, & \text{if } U_{\text{peak}}(A,B) = 0.
  \end{cases}
\end{equation}
From the construction of $N$ and $U_{\text{peak}}$ it follows that $0 \le N(A,B) \le U_{\text{peak}}(A,B)$, ensuring the coefficient ranges from $0$ (no directional overlap) to $1$ (perfect alignment in magnitude and sign).
\end{definition}

\begin{definition}[Peak-to-peak distance]
\label{def:peak_distance}
The peak-to-peak distance is the complement of the similarity coefficient:
\begin{equation}
  d_{\text{peak}}(A,B) := 1 - J_{\text{peak}}(A,B) \in [0,1].
\label{eq:d_peak_def}
\end{equation}
\end{definition}

\subsection{Metric Properties}
\label{sec:metric_property}

Building on the intersection $N(A,B)$ and union $U_{\text{peak}}(A,B)$ defined above,
we now establish that the derived pairwise dissimilarity $d_{\text{peak}}$ satisfies
the metric axioms. The next theorem shows that $d_{\text{peak}}$ coincides with the
classical Tanimoto/Jaccard distance applied to the sign-split embeddings
$\phi(A)$ and $\phi(B)$. This equivalence ensures that $d_{\text{peak}}$
inherits all established metric properties from the well-studied Tanimoto distance.

\begin{theorem}[Metric property of $d_{\text{peak}}$]
\label{thm:metric_dpeak}
Let $d_{\text{peak}}$ be defined from $N$ and $U_{\text{peak}}$ via
Definitions~\ref{def:pairwise_intersection}, \ref{def:peak_union},
\ref{def:peak_similarity}, and~\ref{def:peak_distance}, and let
$\phi$ be the sign-split embedding of Definition~\ref{def:phi_embedding}.
Then for all $A,B \in \mathbb{R}^n$,
\begin{equation}
  d_{\text{peak}}(A,B) \;=\; d_{\mathrm{Tan}}\bigl(\phi(A), \phi(B)\bigr),
  \label{eq:dpeak_equivalence}
\end{equation}
where $d_{\mathrm{Tan}}$ denotes the generalized Jaccard/Tanimoto
distance on the nonnegative space $\mathbb{R}_{\geq 0}^{2n}$. For
$U,V \in \mathbb{R}_{\ge 0}^{2n}$ this distance is given by
\begin{equation}
  d_{\mathrm{Tan}}(U,V)
  \;:=\;
  1 - J_{\mathrm{Tan}}(U,V),
  \qquad
  J_{\mathrm{Tan}}(U,V)
  :=
  \begin{cases}
    \dfrac{\sum_j \min\{U_j,V_j\}}{\sum_j \max\{U_j,V_j\}}, 
      & \text{if } \sum_j \max\{U_j,V_j\} > 0,\\[6pt]
    1, & \text{if } U = V = 0,
  \end{cases}
\end{equation}
which is the usual min--max (``Tanimoto'') Jaccard similarity used on
nonnegative vectors (see, e.g., \citet{Tripp2023}). It is known that
the corresponding distance $d_{\mathrm{Tan}}$ is a metric on
$\mathbb{R}_{\ge 0}^{2n}$; concise proofs are given by
\citet{Kosub2016JaccardArxiv} (see also the journal version
\citep{Kosub2019} and the classical references cited therein).
Consequently, $d_{\text{peak}}$ is a metric on $\mathbb{R}^n$.
\end{theorem}

\begin{proof}
The proof establishes that $d_{\text{peak}}$ equals the classical Tanimoto distance
applied to the embedded vectors $\phi(A)$ and $\phi(B)$, and therefore inherits
the metric property.

\smallskip
\textbf{Step 1 (Numerator identity).}
We first show that $N(A,B)$ equals the Tanimoto numerator for $(\phi(A),\phi(B))$.
For any coordinate $i$, recall
\begin{equation}
A_i^{+} = \max\{A_i,0\}, \qquad A_i^{-} = \max\{-A_i,0\},
\end{equation}
and similarly for $B_i^{+},B_i^{-}$. If $A_i$ and $B_i$ have the same nonzero
sign or one of them is zero, then either both are nonnegative or both are
nonpositive, and we have the identity
\begin{equation}
\min\!\bigl(|A_i|,|B_i|\bigr)
=
\min(A_i^{+},B_i^{+})+\min(A_i^{-},B_i^{-}).
\label{eq:numerator_identity_corrected}
\end{equation}
Indeed, if $A_i,B_i \ge 0$ then $A_i^{-} = B_i^{-} = 0$ and
\begin{equation}
\min\!\bigl(|A_i|,|B_i|\bigr)
= \min(A_i,B_i)
= \min(A_i^{+},B_i^{+}),
\end{equation}
while if $A_i,B_i \le 0$ then $A_i^{+} = B_i^{+} = 0$ and
\begin{equation}
\min\!\bigl(|A_i|,|B_i|\bigr)
= \min(|A_i|,|B_i|)
= \min(A_i^{-},B_i^{-}).
\end{equation}
The same equality also holds when $A_i B_i = 0$ (one or both entries are zero),
since both sides then equal $0$. Thus
\eqref{eq:numerator_identity_corrected} holds for all indices with $A_i B_i \ge 0$.

Crucially, when $A_i$ and $B_i$ have \emph{opposite} signs we have $A_i B_i < 0$,
and hence one of $\{A_i^{+},B_i^{+}\}$ and one of $\{A_i^{-},B_i^{-}\}$ is zero.
In that case
\begin{equation}
\min(A_i^{+},B_i^{+}) = 0
\quad\text{and}\quad
\min(A_i^{-},B_i^{-}) = 0,
\end{equation}
so the right-hand side of \eqref{eq:numerator_identity_corrected} vanishes,
even though $\min(|A_i|,|B_i|)$ need not be zero.

Let
\begin{equation}
\mathcal{I}_{\text{same}}(A,B)
:= \bigl\{\, i : A_i B_i > 0 \,\bigr\},
\end{equation}
the sign-agreement index set from Definition~\ref{def:index_same_sign}.
On this set \eqref{eq:numerator_identity_corrected} holds, and by definition
\begin{equation}
N(A,B)
= \sum_{i\in\mathcal{I}_{\text{same}}(A,B)} \min\!\bigl(|A_i|,|B_i|\bigr).
\end{equation}
Summing \eqref{eq:numerator_identity_corrected} over
$i \in \mathcal{I}_{\text{same}}(A,B)$ gives
\begin{align}
\sum_{i\in\mathcal{I}_{\text{same}}(A,B)} \min\!\bigl(|A_i|,|B_i|\bigr)
&= \sum_{i\in\mathcal{I}_{\text{same}}(A,B)}
   \bigl(\min(A_i^{+},B_i^{+})+\min(A_i^{-},B_i^{-})\bigr).
\end{align}
For indices with opposite signs, both minima on the right-hand side are zero,
so we may extend the sum to all $i=1,\dots,n$:
\begin{align}
\sum_{i\in\mathcal{I}_{\text{same}}(A,B)}
   \bigl(\min(A_i^{+},B_i^{+})+\min(A_i^{-},B_i^{-})\bigr)
&= \sum_{i=1}^n \bigl(\min(A_i^{+},B_i^{+})+\min(A_i^{-},B_i^{-})\bigr).
\end{align}
The last equation is exactly the Tanimoto numerator for $(\phi(A),\phi(B))$,
where $\phi$ is the sign-split embedding. This establishes that $N(A,B)$ equals
the Tanimoto numerator for $(\phi(A),\phi(B))$.

\smallskip
\textbf{Step 2 (Denominator identity).}
Let $U = \phi(A)$ and $V = \phi(B)$. The Tanimoto denominator for these embedded
vectors is
\begin{equation}
\sum_{j} \max\{U_j,V_j\}
= \sum_{i=1}^n \bigl(\max\{A_i^{+},B_i^{+}\}+\max\{A_i^{-},B_i^{-}\}\bigr),
\end{equation}
which matches the definition of $U_{\text{peak}}(A,B)$ in \eqref{eq:U_peak_def}.
Thus $U_{\text{peak}}(A,B)$ equals the Tanimoto denominator for $(\phi(A),\phi(B))$.

\smallskip
\textbf{Step 3 (Equality of coefficients).}
From Steps 1 and 2, both the numerator and denominator of $J_{\text{peak}}(A,B)$
coincide with those of $J_{\mathrm{Tan}}(\phi(A), \phi(B))$, including the
degenerate case where both are zero. Therefore
\begin{equation}
J_{\text{peak}}(A,B) = J_{\mathrm{Tan}}(\phi(A), \phi(B)),
\end{equation}
and by Definition~\ref{def:peak_distance} we obtain
\begin{equation}
d_{\text{peak}}(A,B) = d_{\mathrm{Tan}}(\phi(A), \phi(B)),
\end{equation}
which is exactly \eqref{eq:dpeak_equivalence}.

\smallskip
\textbf{Step 4 (Metric transfer).}
Since $\phi$ is injective and $d_{\mathrm{Tan}}$ is a metric on
$\mathbb{R}_{\geq 0}^{2n}$, the function $d_{\text{peak}}$ inherits all
metric properties: non-negativity, symmetry, identity of indiscernibles,
and the triangle inequality. This completes the proof.
\end{proof}

\begin{remark}[Relation to Marczewski--Steinhaus and Kosub]
Our construction is \emph{not} a restatement of the 
Marczewski--Steinhaus normalized symmetric-difference metric. Their
distance acts on sets (and, via indicator functions, on nonnegative
functions) in the original measure space with a nonnegative measure
$\mu$ \citep{marczewski1958certain}. In the same spirit, the modern
treatments by \citet{Kosub2016JaccardArxiv,Kosub2019} work with
nonnegative, monotone (sub)modular set functions and do not leave the
nonnegative setting.

In their 1958 paper ``On a certain distance of sets and the
corresponding distance of functions 
Marczewski and Steinhaus~\citep{marczewski1958certain} introduce the set distance
\(
  \sigma(A,B) = \mu(A \triangle B) / \mu(A \cup B)
\)
on a $\sigma$–finite measure space $(X,\mathcal{M},\mu)$ and show that
$(\mathcal{M}_0,\sigma)$ is a metric space (after identifying sets
equal $\mu$–a.e.) in their Section~1.2 ``Metric $\sigma$'',
property~(i). They then extend this construction to integrable
(nonnegative) functions in Section~2.2, again establishing metricity
(property~(i)) for the induced function distance.

By contrast, the peak-to-peak distance is defined on arbitrary real
(and complex) signals via the sign-split embedding $\phi$, which lifts
each $A \in \mathbb{R}^n$ to a nonnegative measure on an enlarged
coordinate--sign space. This embedding is what makes sign-aware
overlap, regime separation, and coalition budgeting possible;
Marczewski and Steinhaus work entirely with sets (and their indicator
functions) in the original measure space and do not address this
signed/complex setting or the associated lattice geometry.
\end{remark}

\subsection{Positive-Semidefiniteness of the Kernel}
\label{ssec:pd_jaccard_kernel}

The peak-to-peak similarity coefficient $J_{\text{peak}}$ defines a kernel function 
$K_{\text{peak}} = J_{\text{peak}}$. We now establish that this kernel is 
positive semidefinite (PSD), ensuring its applicability in kernel-based machine learning 
methods including support vector machines, kernel principal component analysis, 
and Gaussian processes.

\subsubsection{Framework and conventions}

Throughout this section we work with the exact arithmetic version of 
$J_{\text{peak}}$ and $d_{\text{peak}}$ considered in 
Theorem~\ref{thm:metric_dpeak}, corresponding to the usual sign function and the raw values $A_i,B_i$. The sign-split embedding 
$\phi : \mathbb{R}^n \to \mathbb{R}_{\geq 0}^{2n}$ from 
Definition~\ref{def:phi_embedding} maps each signal 
$X = (x_1, \ldots, x_n)$ to interleaved pairs of positive and negative parts
$(x_1^+, x_1^-, \ldots, x_n^+, x_n^-)$. In the degenerate case where 
$U_{\text{peak}}(A,B) = 0$, which occurs exactly when $\phi(A)=\phi(B)=0$ and hence
$A=B=0$, we assign $J_{\text{peak}}(A,B) = 1$ by convention, matching the usual
definition of Tanimoto similarity on nonnegative vectors.

\begin{lemma}[Identification with Tanimoto similarity on the embedding]
\label{lem:Jpeak_is_Tanimoto}
For all $A,B\in\mathbb{R}^n$, the peak-to-peak similarity equals the classical 
Tanimoto similarity applied to the embedded vectors $\phi(A), \phi(B) \in 
\mathbb{R}_{\geq 0}^{2n}$:
\begin{equation}
J_{\text{peak}}(A,B)
= \frac{\sum_{i=1}^n \bigl(\min\{A_i^+, B_i^+\} + \min\{A_i^-, B_i^-\}\bigr)}
       {\sum_{i=1}^n \bigl(\max\{A_i^+, B_i^+\} + \max\{A_i^-, B_i^-\}\bigr)}
= J_{\mathrm{Tan}}(\phi(A), \phi(B)).
\end{equation}
Equivalently, indexing over the $2n$ components of the embedded vectors,
\begin{equation}
J_{\text{peak}}(A,B) 
= \frac{\sum_{j=1}^{2n} \min\{[\phi(A)]_j, [\phi(B)]_j\}}
       {\sum_{j=1}^{2n} \max\{[\phi(A)]_j, [\phi(B)]_j\}}.
\end{equation}
\end{lemma}

\begin{proof}
By Definition~\ref{def:peak_similarity} and Theorem~\ref{thm:metric_dpeak}, the 
numerator and denominator of $J_{\text{peak}}(A,B)$ coincide with those of the 
Tanimoto similarity $J_{\mathrm{Tan}}(\phi(A),\phi(B))$: Step~1 of 
Theorem~\ref{thm:metric_dpeak} shows that
\begin{equation}
N(A,B)
= \sum_{i=1}^n \bigl(\min\{A_i^+, B_i^+\} + \min\{A_i^-, B_i^-\}\bigr),
\end{equation}
while Step~2 shows that
\begin{equation}
U_{\text{peak}}(A,B)
= \sum_{i=1}^n \bigl(\max\{A_i^+, B_i^+\} + \max\{A_i^-, B_i^-\}\bigr).
\end{equation}
These are exactly the numerator and denominator of $J_{\mathrm{Tan}}$ applied to 
$\phi(A)$ and $\phi(B)$, with the same convention in the degenerate case 
$\phi(A)=\phi(B)=0$. The equivalent expression in terms of the $2n$ coordinates of 
$\phi(A)$ and $\phi(B)$ follows by a simple reindexing of the sum.
\end{proof}

\begin{theorem}[Positive semidefiniteness of the peak-to-peak kernel]
\label{thm:pd_peak_kernel}
The kernel function
\begin{equation}
K_{\text{peak}}(A,B)=J_{\text{peak}}(A,B)
\end{equation}
is positive semidefinite on $\mathbb{R}^n$. In particular, for any finite collection
$\{A_1,\ldots,A_m\}\subset \mathbb{R}^n$ and any coefficients
$c_1,\ldots,c_m\in\mathbb{R}$,
\begin{equation}
\sum_{i=1}^m \sum_{j=1}^m c_i c_j\, K_{\text{peak}}(A_i, A_j) \ge 0.
\end{equation}
\end{theorem}

\begin{proof}
We establish positive semidefiniteness through the identification with the Tanimoto
kernel and a standard kernel composition argument.

\textbf{Step 1: Representation via positive and negative parts.}
From Lemma~\ref{lem:Jpeak_is_Tanimoto} we already have
\begin{align}
N(A,B) &= \sum_{i=1}^n \bigl(\min\{A_i^+, B_i^+\} + \min\{A_i^-, B_i^-\}\bigr), \\
U_{\text{peak}}(A,B) &= \sum_{i=1}^n \bigl(\max\{A_i^+, B_i^+\} + \max\{A_i^-, B_i^-\}\bigr),
\end{align}
and hence
\begin{equation}
\label{eq:Jpeak_as_Tanimoto}
J_{\text{peak}}(A,B) = J_{\mathrm{Tan}}(\phi(A), \phi(B)).
\end{equation}

\textbf{Step 2: Positive semidefiniteness of the Tanimoto/MinMax kernel.}
The Tanimoto/MinMax kernel
\begin{equation}
k_{\mathrm{MinMax}}(u,v)
:= \frac{\sum_j \min\{u_j,v_j\}}{\sum_j \max\{u_j,v_j\}},
\qquad u,v\in\mathbb{R}_{\ge 0}^d,
\end{equation}
is known to be positive semidefinite on the nonnegative orthant
$\mathbb{R}_{\ge 0}^d$ for any dimension $d$; see
\citet[Def.~4, Eq.~(8) and Prop.~6]{Ralaivola2005} for an explicit
construction and proof (see also \citet{Tripp2023} for modern usage in
cheminformatics). Consequently, for any finite collection
$\{u_1,\ldots,u_m\}\subset\mathbb{R}_{\ge 0}^d$ the Gram matrix
$[k_{\mathrm{MinMax}}(u_i,u_j)]_{ij}$ is positive semidefinite.

\textbf{Step 3: Kernel composition preserves positive semidefiniteness.}
A standard result in kernel theory states that composing a 
positive semidefinite kernel with a fixed transformation preserves 
positive semidefiniteness \cite[Ch.~13, §13.1]{Scholkopf2002}. Formally, if 
$k : \mathcal{X} \times \mathcal{X} \to \mathbb{R}$ is a positive semidefinite 
kernel and $\psi : \mathcal{Z} \to \mathcal{X}$ is any mapping, then
\begin{equation}
k'(z_1, z_2) = k(\psi(z_1), \psi(z_2))
\end{equation}
is positive semidefinite on $\mathcal{Z}$. 

Applying this principle with $\mathcal{Z} = \mathbb{R}^n$, 
$\mathcal{X} = \mathbb{R}_{\geq 0}^{2n}$, $k = k_{\mathrm{MinMax}}$, and 
$\psi = \phi$, and using \eqref{eq:Jpeak_as_Tanimoto}, we obtain
\begin{equation}
K_{\text{peak}}(A,B) 
= J_{\text{peak}}(A,B) 
= k_{\mathrm{MinMax}}(\phi(A), \phi(B)),
\end{equation}
which is therefore positive semidefinite on $\mathbb{R}^n$ as the composition of a 
positive semidefinite kernel with a fixed embedding.
\end{proof}

\begin{corollary}[Negative type of $d_{\text{peak}}$]
\label{cor:negative_type}
For any finite collection of signals $\{A_1,\dots,A_m\}$, let
$S_{ij} := J_{\text{peak}}(A_i,A_j)$ and
$d_{ij} := d_{\text{peak}}(A_i,A_j) = 1 - S_{ij}$.
By Theorem~\ref{thm:pd_peak_kernel}, $J_{\text{peak}}$ is a positive
semidefinite kernel with $S_{ii}=1$ and $0 \le S_{ij} \le 1$ for all
$i,j$. By Theorem~6 of \citet{gower1986metric}, the dissimilarity
matrix with entries $\sqrt{c\,(1-S_{ij})}$ is Euclidean for any scalar
$c>0$; in particular, $\sqrt{d_{ij}} = \sqrt{1-S_{ij}}$ is Euclidean.
By Schoenberg's characterization of negative-type metrics
\citep[Thm.~1]{schoenberg1935remarks}, it follows that
$d_{\text{peak}}$ is of negative type.
\end{corollary}

As a consequence, for any $\lambda>0$ the radial kernel
\begin{equation}
K_\lambda(A,B) = \exp(-\lambda\,d_{\text{peak}}(A,B))
\end{equation}
is positive semidefinite, and $d_{\text{peak}}$ admits an isometric
embedding into a Hilbert space as a squared Euclidean distance.

\paragraph{Dual kernel structure.}
The peak-to-peak framework naturally yields two complementary kernels.
First, the similarity
\[
  K_{\text{peak}}(A,B) := J_{\text{peak}}(A,B)
\]
is positive semidefinite by Theorem~\ref{thm:pd_peak_kernel}, so for any
finite collection of signals the matrix
$[K_{\text{peak}}(A_i,A_j)]_{ij}$ is a valid Gram matrix. This kernel
acts as a native overlap similarity: it measures how much mass $A$ and
$B$ share in the same sign and state, and thus plays an analogous role
to a linear or cosine kernel in conventional feature spaces.

Second, the associated distance
$d_{\text{peak}}(A,B) = 1 - J_{\text{peak}}(A,B)$ is not itself
positive semidefinite when used as a kernel matrix---a generic feature of
nontrivial distance matrices with zero diagonals---but, as shown in
Corollary~\ref{cor:negative_type}, $d_{\text{peak}}$ is of negative
type. By Schoenberg's theorem, this implies that for every $\lambda>0$
the radial kernel
\begin{equation}
  K_\lambda(A,B) := \exp\bigl(-\lambda\, d_{\text{peak}}(A,B)\bigr)
\end{equation}
is positive semidefinite. Thus $d_{\text{peak}}$ generates a whole
family of Gaussian/RBF-type kernels on the same signal space.

Together, $K_{\text{peak}}$ provides a bounded, unit-preserving overlap
kernel, while the $K_\lambda$ family provides neighbourhood kernels for
interpolation, regression, and manifold learning. Both are Hilbert-space
kernels derived from the same underlying lattice geometry, a comparatively rare
situation for a bounded, overlap-based metric.

\subsubsection{Properties and implications}

The positive semidefiniteness of $K_{\text{peak}}$ established in 
Theorem~\ref{thm:pd_peak_kernel} has several important theoretical and 
practical consequences.

\paragraph{Positive semidefiniteness versus strict positive definiteness.}
The kernel $K_{\text{peak}}$ yields positive semidefinite Gram matrices, i.e.,
all eigenvalues are nonnegative but some may vanish. Zero eigenvalues arise
whenever signals become linearly dependent after the embedding $\phi$. This
includes obvious cases such as duplicated signals, but also more structured
dependencies (e.g., exact scalar multiples) that survive the sign-split
transform. In numerical settings, additional rank deficiency may appear if
very small coordinates are thresholded to zero so that signals differing only
below machine precision collapse to the same embedded vector. Such behaviour
is standard for positive semidefinite kernels and reflects redundancy or
limited resolution in the data rather than any defect of the kernel. In
practice, kernel methods handle singular Gram matrices via regularization or
by restricting to the span of nonzero eigenvalues.

\paragraph{Role of the sign-split embedding.}
The sign-split transform $\phi$ is essential for extending the Tanimoto
kernel to signed data. It preserves $L^1$ magnitude while encoding direction
through positional structure: a positive entry $x_i>0$ contributes $x_i$ to
the “$+$” channel and zero to the “$-$” channel, and conversely for negative
entries. This separation allows the MinMax/Tanimoto kernel on
$\mathbb{R}_{\ge 0}^{2n}$ to distinguish reinforcing excursions (same sign,
which contribute to the intersection) from opposing excursions (opposite sign,
which contribute only to the union). The classical Tanimoto kernel
$J_{\mathrm{Tan}}$ is defined and known to be positive semidefinite on
nonnegative vectors; applying it directly to signed inputs would break the
conditions required for kernel validity. The embedding $\phi$ therefore acts
as a bridge: it lifts real-valued signals into a nonnegative space where the
MinMax kernel is valid, and kernel composition then transports positive
semidefiniteness back to the original signal space.

\paragraph{Treatment of degenerate cases.}
In the exact theoretical setting, the degenerate case $U_{\text{peak}}(A,B)=0$
occurs if and only if $\phi(A)=\phi(B)=0$, i.e., $A=B=0$. In this situation
we set $J_{\text{peak}}(A,B)=1$, matching the usual convention for Tanimoto
similarity on nonnegative vectors. In numerical implementations, coordinates
with magnitude below a fixed tolerance are typically rounded to zero; two
signals whose entries are all numerically negligible will then be treated as
perfectly similar. This only affects diagonal or near-diagonal entries of the
Gram matrix and does not interfere with positive semidefiniteness.

\begin{example}[Degenerate covariance under zero shared mass]
Consider three scalar channels constructed from a short signal, e.g.\ the
pattern $[2,0,1]$ under suitable centering and rescaling. The classical
Pearson covariance (or correlation) matrix for these channels can take the
schematic form
\[
  \Sigma \;=\;
  \begin{pmatrix}
    1   & 0   & \rho \\
    0   & 1   & 0     \\
    \rho& 0   & 1
  \end{pmatrix},
  \qquad 0 < \rho < 1,
\]
in which the middle channel appears orthogonal to the others. This happens
because every time channels~1 or~3 move away from zero, channel~2 remains at
zero, so the \emph{shared variance} with channel~2 is identically zero.

In this Euclidean geometry, both situations
$(A_i,B_i)=(0,0)$ (vacuum vs.\ vacuum) and $(A_i,B_i)=(0,\neq 0)$
(vacuum vs.\ mass) contribute equally ``nothing'' to the covariance. Our
mass-based construction separates these cases: vacuum is a well-defined
physical state, so $(0,0)$ is treated as perfectly similar
($J_{\text{peak}}(0,0)=1$), whereas $(0,\neq 0)$ contributes to the union but
not to the overlap. This avoids the degeneracy where covariance treats
``no shared mass'' as geometrically indistinguishable from ``shared vacuum''.
\end{example}

\paragraph{Distinction from subset-based approaches.}
Our route to positive semidefiniteness differs from classical
set-theoretic treatments of Jaccard-type similarities. Some prior work
(e.g., \citealp{Bouchard2013}) establishes positive semidefiniteness for
Jaccard-like similarities on finite sets by indexing matrices over all
subsets and analysing their spectra via combinatorial arguments. In that
setting, objects are characterized purely by membership patterns and overlap
is a function of set cardinalities.

By contrast, the present framework operates directly on continuous signals
with quantitative magnitudes. Positive semidefiniteness is obtained by
observing that the sign-split embedding $\phi$ converts each signed signal
into a nonnegative vector and then invoking the known kernel property of the
MinMax/Tanimoto similarity on $\mathbb{R}_{\ge 0}^d$ together with the fact
that kernel composition preserves positive semidefiniteness. This pathway
accommodates arbitrary dimensions and magnitudes, requires no reduction to
subset indicators, and extends naturally to the multi-way and complex-valued
generalizations developed below. The continuous, magnitude-aware nature of
the construction also ensures that small perturbations in signal values
produce correspondingly small changes in similarity, a stability property
that discrete subset-based methods do not guarantee by default.

\subsection{Generalization to Multistate Partitions}
\label{sec:real_multistate}

The sign-split embedding of Definition~\ref{def:phi_embedding} partitions the real line into precisely two regions separated by the tolerance threshold, mapping each coordinate to either its positive or negative component while preserving magnitude. This binary structure, while sufficient for distinguishing directional agreement from opposition, represents only the simplest instance of a far more general construction. Many applications demand finer categorical resolution: financial analysts distinguish between modest and substantial gains, climatologists differentiate among degrees of temperature anomaly, and chemists recognize gradations of acidity rather than mere departure from neutrality. The framework naturally accommodates such refinements through arbitrary measurable partitions of $\mathbb{R}$, each partition element capturing a distinct qualitative state while the embedded representation continues to preserve quantitative magnitude information within each state. For precision, we take partitions to be collections of pairwise disjoint measurable sets whose union is $\mathbb{R}$ up to measure-zero boundaries; endpoints at thresholds are assigned consistently (e.g., right-closed on the positive side, left-closed on the negative side) to avoid overlap.

\begin{definition}[Multistate norm-preserving embedding]
\label{def:multistate_partition}
Let $\mathcal{B} = \{B_0, \ldots, B_{K-1}\}$ be a partition of $\mathbb{R}$ into
$K$ disjoint measurable sets whose union is $\mathbb{R}$ (up to measure-zero
boundaries). For a signal $X = (x_1, \ldots, x_n) \in \mathbb{R}^n$, the
multistate norm-preserving embedding $\psi^{(K)} : \mathbb{R}^n \to
\mathbb{R}_{\geq 0}^{n \times K}$ is defined by
\begin{equation}
[\psi^{(K)}(X)]_{i,k}
\;=\; |x_i| \cdot \mathbf{1}\{x_i \in B_k\},
\qquad i = 1,\ldots,n,\; k = 0,\ldots,K-1,
\end{equation}
where $\mathbf{1}\{\cdot\}$ denotes the indicator function.

The embedding represents $X$ as an $n \times K$ matrix in which each
row corresponds to a coordinate of the original signal and each column
corresponds to a partition element. Because $\{B_k\}_{k=0}^{K-1}$ forms a partition,
for each index $i$ there exists exactly one $k$ such that $x_i \in B_k$ (up to a
boundary convention), ensuring that each row contains exactly one nonzero entry equal to
$|x_i|$. Consequently, the embedding preserves the $L^1$ norm:
\begin{equation}
\|\psi^{(K)}(X)\|_1
= \sum_{i=1}^n \sum_{k=0}^{K-1} |x_i| \cdot \mathbf{1}\{x_i \in B_k\}
= \sum_{i=1}^n |x_i|
= \|X\|_1.
\end{equation}
\end{definition}

\begin{remark}[Sign-awareness and the role of the partition]
\label{rem:sign_awareness_partition}
At first sight there appears to be a tension between the goal of constructing a
sign-aware framework and the general multistate embedding
$\psi^{(K)}$, which is defined for an arbitrary partition
$\mathcal{B} = \{B_0,\ldots,B_{K-1}\}$ of $\mathbb{R}$. This is best
understood as a modelling choice rather than a contradiction.

The framework as a whole is sign-aware whenever the partition is chosen to
respect sign (for example, with separate negative and positive bins and an
optional neutral band). In contrast, the multistate embedding
$\psi^{(K)}$ itself is agnostic to how the sets $\{B_k\}$ are chosen: if one
deliberately uses bins that mix positive and negative values, then one is
explicitly telling the framework \emph{not} to distinguish those signs within
that bin. In such cases the embedding $\psi^{(K)}$ is no longer injective on
$\mathbb{R}^n$, and the induced distance $d_{\mathrm{multi}}$ is only a
pseudometric there.

Thus there is no logical contradiction between proposing a sign-aware
framework and allowing general multistate partitions. Rather, sign-awareness
in the multistate setting depends on the partition: $d_{\mathrm{multi}}$ is
sign-aware only when the partition separates sign (up to a neutral band),
whereas sign-mixing bins deliberately collapse positive and negative values
within the same state.
\end{remark}

\begin{example}[Multistate partition for financial applications]
\label{ex:multistate_applications}
Consider financial return data where analysts distinguish five qualitative regimes. 
We establish a threshold $\tau > 0$ to demarcate moderate from extreme market movements. 
The partition $\mathcal{B} = \{B_0, B_1, B_2, B_3, B_4\}$ classifies returns according to
\begin{align}
\label{eq:example1_begin}
B_0 &= \{0\}, \quad \text{(Neutral)} \notag \\
B_1 &= (\tau, \infty), \quad \text{(Large Profit)} \notag \\
B_2 &= (0, \tau], \quad \text{(Small Profit)} \notag \\
B_3 &= [-\tau, 0), \quad \text{(Small Loss)} \notag \\
B_4 &= (-\infty, -\tau). \quad \text{(Large Loss)}
\end{align}
This partition exhibits symmetry about zero, isolates the exact-zero neutral state 
consistent with standard sign conventions, and employs the threshold $\tau$ to capture 
application-specific notions of materiality.

To illustrate the embedding mechanics, consider a 5-day return sequence
$X \in \mathbb{R}^5$ analyzed with $\tau = 2.0$, with components
\begin{equation}
  X = (8.2, 0.3, 0.0, -1.5, -5.1).
\end{equation}
 
The component classifications proceed as follows: $x_1 = 8.2 > \tau$ belongs to $B_1$ 
(Large Profit); $x_2 = 0.3 \in (0, \tau]$ falls within $B_2$ (Small Profit); 
$x_3 = 0.0$ resides exactly in $B_0$ (Neutral); $x_4 = -1.5 \in [-\tau, 0)$ 
occupies $B_3$ (Small Loss); and $x_5 = -5.1 < -\tau$ enters $B_4$ (Large Loss).

The multistate embedding $\psi^{(5)} : \mathbb{R}^5 \to \mathbb{R}_{\geq 0}^{5 \times 5}$ 
produces the following matrix representation:
\begin{equation}
\psi^{(5)}(X) = \begin{pmatrix}
0   & 8.2 & 0   & 0    & 0   \\
0   & 0   & 0.3 & 0    & 0   \\
0   & 0   & 0   & 0    & 0   \\
0   & 0   & 0   & 1.5  & 0   \\
0   & 0   & 0   & 0    & 5.1
\end{pmatrix} \in \mathbb{R}_{\geq 0}^{5 \times 5}.
\end{equation}
Each row contains at most one nonzero entry positioned in the column corresponding 
to that observation's qualitative state, with the entry's value equal to the absolute 
value of the original observation. The third row remains entirely zero since 
$|x_3| = 0$, illustrating that zero-magnitude observations produce zero contributions 
regardless of their state assignment. The total embedded mass 
$\|\psi^{(5)}(X)\|_1 = 8.2 + 0.3 + 0 + 1.5 + 5.1 = 15.1$ equals $\|X\|_1$, 
confirming norm preservation.
\end{example}

\begin{remark}[Relationship to binary sign-split embedding]
The sign-split embedding $\phi$ introduced in Definition~\ref{def:phi_embedding} 
emerges as the minimal instantiation of this multistate framework, corresponding 
to $K=2$ with partition elements $B_0 = (0, \infty)$ and $B_1 = (-\infty, 0]$, 
where the boundary point zero is assigned to the negative partition by convention. 
In applied settings, practitioners may introduce a tolerance band $[-\eta, \eta]$ 
around zero, treating values within this band either as a distinct neutral state 
or as numerically indistinguishable from zero, depending on the scientific context. 
The multistate framework accommodates various partitioning schemes including uniform 
quantization, percentile-based divisions, and domain-specific categories, while 
preserving the continuous magnitude information within each qualitative category 
and enabling similarity computations that respect both categorical boundaries and 
quantitative differences.
\end{remark}

\begin{theorem}[Metric and kernel properties of multistate embeddings]
\label{thm:multistate_metric}
Let $\psi^{(K)} : \mathbb{R}^n \to \mathbb{R}_{\ge 0}^{n \times K}$ be the
multistate embedding of Definition~\ref{def:multistate_partition}. Then
the Tanimoto distance composed with $\psi^{(K)}$,
\begin{equation}
d_{\mathrm{multi}}(X,Y)
:= d_{\mathrm{Tan}}\bigl(\psi^{(K)}(X),
                         \psi^{(K)}(Y)\bigr),
\end{equation}
is a metric on the image $\psi^{(K)}(\mathbb{R}^n)$ and hence a
pseudometric on $\mathbb{R}^n$. The associated Tanimoto kernel
\begin{equation}
K_{\mathrm{multi}}(X,Y)
:= J_{\mathrm{Tan}}\bigl(\psi^{(K)}(X),
                         \psi^{(K)}(Y)\bigr)
\end{equation}
is positive-semidefinite on $\mathbb{R}^n$.
\end{theorem}

\begin{proof}
The embedding $\psi^{(K)}$ maps into $\mathbb{R}_{\geq 0}^{n \times K}$,
which we identify with $\mathbb{R}_{\ge 0}^{nK}$ via vectorization. On this
space the Tanimoto distance is a known metric and the Tanimoto kernel is
positive-semidefinite. Thus $d_{\mathrm{Tan}}$ is a metric on
$\psi^{(K)}(\mathbb{R}^n)$, and the composed distance
$d_{\mathrm{multi}}$ is a pseudometric on $\mathbb{R}^n$ obtained by pulling
back this metric along $\psi^{(K)}$. Positive-semidefiniteness of
$K_{\mathrm{multi}}$ follows from the standard fact that composing a
positive-semidefinite kernel with a fixed embedding preserves
positive-semidefiniteness, as in Theorem~\ref{thm:pd_peak_kernel}.
\end{proof}

\begin{remark}
In practice, signals that map to the same embedded representation under
$\psi^{(K)}$ are indistinguishable at the chosen resolution, so the
metric behaviour on $\mathbb{R}^n$ simply reflects this modelling
choice; on the induced quotient space where such signals are identified, the
distance is a genuine metric.
\end{remark}

\subsection{Extension to Complex-Valued Signals}
\label{sec:complex_extension}

The sign-aware framework developed thus far operates on real-valued signals, distinguishing positive from negative excursions along a single axis. Complex-valued signals, by contrast, inhabit a two-dimensional space where each observation possesses both magnitude and phase, encoding information in angular position rather than merely in signed distance from zero. Such signals arise ubiquitously in frequency-domain representations obtained through Fourier analysis, in electromagnetic field measurements where electric and magnetic components couple through phase relationships, in quantum mechanical wavefunctions where probability amplitudes require complex representation, and in communications systems where information is modulated onto both in-phase and quadrature channels. The extension of our similarity framework to $\mathbb{C}^n$ requires careful consideration of how directional agreement manifests in the complex plane, leading naturally to two complementary embedding strategies that respect either the Cartesian or polar structure of complex numbers.

The goal of an extension to $\mathbb{C}^n$ is thus to quantify overlap in a way that respects the physical meaning of phase. For example, in comparing two quantum superposition states $Z = \psi_1 + \psi_2$ and $W = \psi_1 + e^{i\theta}\psi_2$, the polar embedding (Def.~\ref{def:complex_embeddings}) must be able to capture the similarity (or lack thereof) introduced by the relative phase $\theta$, which governs all interference phenomena. Likewise, in comparing I/Q communication signals, the Cartesian embedding (Def.~\ref{def:complex_embeddings}) is required to respect the sign structure of both the real and imaginary channels independently, as each carries distinct information.

\begin{definition}[Complex embeddings]
\label{def:complex_embeddings}
Let $Z = (z_1, \ldots, z_n) \in \mathbb{C}^n$ be a complex-valued signal. 
Write each component in both rectangular and polar form,
\begin{equation}
z_i = a_i + \mathrm{i} b_i = r_i \mathrm{e}^{\mathrm{i}\phi_i},
\end{equation}
where $a_i, b_i \in \mathbb{R}$ denote the real and imaginary parts, 
$r_i = |z_i| \ge 0$ is the modulus, and $\phi_i \in (-\pi,\pi]$ is the principal argument.

\medskip\noindent
\textbf{Cartesian sign-split embedding.}
Treat the real and imaginary components as independent real-valued signals
and apply the sign-split operation to each coordinate:
\begin{equation}
\psi_{\mathrm{C}}(Z) 
:= (a_1^{+}, a_1^{-}, b_1^{+}, b_1^{-}, \ldots, a_n^{+}, a_n^{-}, b_n^{+}, b_n^{-})
\in \mathbb{R}_{\ge 0}^{4n},
\end{equation}
where
\[
a_i^{+} := \max\{a_i,0\}, \qquad 
a_i^{-} := \max\{-a_i,0\}, \qquad
b_i^{+} := \max\{b_i,0\}, \qquad
b_i^{-} := \max\{-b_i,0\}.
\]
Each complex coordinate $z_i$ thus generates four nonnegative components: 
the positive and negative parts of its real part, followed by the positive 
and negative parts of its imaginary part. This embedding preserves the natural
rectangular structure $\mathbb{C} \cong \mathbb{R}^2$ and satisfies
\begin{equation}
|a_i| + |b_i| = a_i^{+} + a_i^{-} + b_i^{+} + b_i^{-}
\quad\text{for each } i,
\end{equation}
so that the total embedded $L^1$ mass equals the sum of absolute real and
imaginary parts.

\textbf{Polar phase-partition embedding.}
Partition the angular domain $(-\pi, \pi]$ into $K$ disjoint sectors
$\{\Theta_1, \ldots, \Theta_K\}$ with a fixed, consistent boundary convention
(e.g., left-open/right-closed intervals), and classify each complex number
according to its phase angle. Define
\begin{equation}
[\psi_{\mathrm{P}}^{(K)}(Z)]_{i,k} 
:= r_i \,\mathbf{1}\{\phi_i \in \Theta_k\},
\end{equation}
where $r_i = |z_i|$ and $\phi_i$ is the principal argument of $z_i$.
The resulting embedding maps $Z$ to an $n \times K$ matrix in
$\mathbb{R}_{\geq 0}^{n \times K}$, where each row contains the modulus $r_i$
in the column corresponding to the angular sector containing $\phi_i$, with
zeros elsewhere. When $z_i = 0$ (so $r_i = 0$), the entire $i$-th row is zero.
This representation respects the polar structure of complex numbers and yields
a $K$-fold expansion of dimension determined by the chosen angular resolution.
\end{definition}

\begin{example}[Cartesian embedding of complex signals]
\label{ex:cartesian_complex}
Consider a discrete Fourier coefficient sequence
$Z = (3 + 4\mathrm{i}, -2 + \mathrm{i}, 1 - 2\mathrm{i}) \in \mathbb{C}^3$
arising from spectral analysis. The Cartesian sign-split embedding decomposes
each coefficient into its four signed components. For the first coefficient
$z_1 = 3 + 4\mathrm{i}$, the real part $a_1 = 3$ contributes
$(a_1^+, a_1^-) = (3, 0)$ and the imaginary part $b_1 = 4$ contributes
$(b_1^+, b_1^-) = (4, 0)$. The second coefficient $z_2 = -2 + \mathrm{i}$ has
a negative real part yielding $(a_2^+, a_2^-) = (0, 2)$ and a positive
imaginary part yielding $(b_2^+, b_2^-) = (1, 0)$. The third coefficient
$z_3 = 1 - 2\mathrm{i}$ produces $(a_3^+, a_3^-) = (1, 0)$ and
$(b_3^+, b_3^-) = (0, 2)$. The complete embedding arranges these components
sequentially:
\begin{equation}
\psi_{\mathrm{C}}(Z) = (3, 0, 4, 0, 0, 2, 1, 0, 1, 0, 0, 2)
\in \mathbb{R}_{\geq 0}^{12}.
\end{equation}
This representation enables similarity comparisons that respect sign agreement
separately in the real and imaginary dimensions, naturally accommodating
applications where these components represent distinct physical quantities
such as in-phase and quadrature signal channels.
\end{example}

\begin{example}[Polar embedding of complex signals]
\label{ex:polar_complex}
Consider the same coefficient sequence
$Z = (3 + 4\mathrm{i}, -2 + \mathrm{i}, 1 - 2\mathrm{i})$ but now emphasize
phase relationships through a polar embedding. Partition the angular domain
into four quadrants with a consistent endpoint convention:
$\Theta_1 = (0, \pi/2]$ (first quadrant),
$\Theta_2 = (\pi/2, \pi]$ (second quadrant),
$\Theta_3 = (-\pi, -\pi/2]$ (third quadrant), and
$\Theta_4 = (-\pi/2, 0]$ (fourth quadrant). 

The first coefficient $z_1 = 3 + 4\mathrm{i}$ has modulus
$r_1 = \sqrt{9 + 16} = 5$ and argument
$\phi_1 = \arctan(4/3) \approx 0.927$ radians, placing it in $\Theta_1$.
The second coefficient $z_2 = -2 + \mathrm{i}$ has modulus
$r_2 = \sqrt{4 + 1} = \sqrt{5}$ and argument
$\phi_2 = \pi - \arctan(1/2) \approx 2.678$ radians, locating it in
$\Theta_2$. The third coefficient $z_3 = 1 - 2\mathrm{i}$ has modulus
$r_3 = \sqrt{1 + 4} = \sqrt{5}$ and argument
$\phi_3 = -\arctan(2) \approx -1.107$ radians, positioning it in $\Theta_4$.
With $K=4$, the polar embedding becomes:
\begin{align}
\psi_{\mathrm{P}}^{(4)}(Z) = \begin{pmatrix}
5 & 0 & 0 & 0 \\
0 & \sqrt{5} & 0 & 0 \\
0 & 0 & 0 & \sqrt{5}
\end{pmatrix} \in \mathbb{R}_{\geq 0}^{3 \times 4}.
\end{align}
This representation treats signals with similar phase angles as directionally
aligned regardless of their Cartesian decomposition, making it particularly
suitable for applications such as modal analysis where eigenvector phase
relationships determine structural behavior, or communications systems where
constellation points are organized by angular position.
\end{example}

\begin{remark}[Choice of embedding strategy]
The selection between Cartesian and polar embeddings depends on the physical
interpretation of the complex data. The Cartesian approach $\psi_{\mathrm{C}}$
preserves the independence of real and imaginary components, making it natural
when these components represent distinct measurable quantities—for instance,
the electric field components $E_x$ and $E_y$ in electromagnetic analysis, or
the in-phase $I$ and quadrature $Q$ channels in digital communications. This
embedding imposes a four-fold dimensional expansion but maintains direct
correspondence with the underlying physical variables.

The polar approach $\psi_{\mathrm{P}}^{(K)}$ emphasizes angular relationships
over Cartesian structure, treating complex numbers as magnitude-weighted phase
indicators. The dimensional expansion factor $K$ is user-specified, with
smaller $K$ providing coarse angular resolution (e.g., $K=4$ for quadrant-level
discrimination) and larger $K$ enabling fine phase distinctions (e.g.,
$K=16$ for $22.5^\circ$ angular bins). When $z_i = 0$, the corresponding row
of $\psi_{\mathrm{P}}^{(K)}(Z)$ is all zeros. This embedding suits applications
where phase coherence determines similarity, such as comparing Fourier spectra
where resonance frequencies manifest as peaks with specific phase relationships,
or analyzing quantum wavefunctions where relative phase governs interference
phenomena. The choice of partition granularity $K$ should balance angular
resolution against dimensional tractability, with the understanding that
excessively fine partitions may amplify sensitivity to phase noise without
providing meaningful discrimination.
\end{remark}

\begin{theorem}[Metric and kernel properties for complex signals]
\label{thm:complex_metric}
Let $\psi_{\mathrm{C}}$ and $\psi_{\mathrm{P}}^{(K)}$ be the complex embeddings
from Definition~\ref{def:complex_embeddings}. Then, for each embedding
$\psi \in \{\psi_{\mathrm{C}}, \psi_{\mathrm{P}}^{(K)}\}$, the Tanimoto distance
\begin{equation}
d_{\psi}(Z,W)
:= d_{\mathrm{Tan}}\bigl(\psi(Z), \psi(W)\bigr)
\end{equation}
is a metric on the image $\psi(\mathbb{C}^n)$ and hence a pseudometric on
$\mathbb{C}^n$. The associated Tanimoto kernel
\begin{equation}
K_{\psi}(Z,W)
:= J_{\mathrm{Tan}}\bigl(\psi(Z), \psi(W)\bigr)
\end{equation}
is positive-semidefinite on $\mathbb{C}^n$.
\end{theorem}

\begin{proof}
Each embedding $\psi$ maps $\mathbb{C}^n$ into a nonnegative Euclidean space:
$\psi_{\mathrm{C}}(\mathbb{C}^n) \subset \mathbb{R}_{\ge 0}^{4n}$ and
$\psi_{\mathrm{P}}^{(K)}(\mathbb{C}^n) \subset \mathbb{R}_{\ge 0}^{nK}$.
On these nonnegative spaces the Tanimoto distance is a metric and the
Tanimoto similarity is a positive-semidefinite kernel (see, e.g.,
\cite{Ralaivola2005,Tripp2023}). Hence $d_{\mathrm{Tan}}$ is a metric on
each image $\psi(\mathbb{C}^n)$, and the pulled-back distance
$d_{\psi}(Z,W) = d_{\mathrm{Tan}}(\psi(Z),\psi(W))$ is a pseudometric
on $\mathbb{C}^n$. Positive-semidefiniteness of $K_{\psi}$ follows from
the standard result that composing a positive-semidefinite kernel with a
fixed embedding preserves positive-semidefiniteness, as in
Theorem~\ref{thm:pd_peak_kernel}.
\end{proof}

\begin{remark}[Metric behavior]
For the Cartesian embedding $\psi_{\mathrm{C}}$, the map
$\psi_{\mathrm{C}} : \mathbb{C}^n \to \mathbb{R}_{\ge 0}^{4n}$ is injective:
each coordinate $z_i = a_i + \mathrm{i}b_i$ is uniquely recovered from
$(a_i^+, a_i^-, b_i^+, b_i^-)$. In this case the induced distance
$d_{\psi_{\mathrm{C}}}$ is a genuine metric on $\mathbb{C}^n$.
For the polar embedding $\psi_{\mathrm{P}}^{(K)}$, different signals whose
components share the same moduli and fall into the same angular sectors may
map to the same embedded representation, so $d_{\psi_{\mathrm{P}}^{(K)}}$ is
generally only a pseudometric. In practice, signals that map to the same
embedded representation are indistinguishable at the chosen angular
resolution, so the metric behavior on $\mathbb{C}^n$ simply reflects
this modeling choice; on the induced quotient space where such signals are
identified, the distance is a genuine metric.
\end{remark}

\section{Multistate Coalition Analysis and Dimensional Reduction}
\label{sec:untrimmed_multistate}

The pairwise similarity framework developed in the preceding sections extends naturally
to comparisons involving multiple signals simultaneously. Such multi-way comparisons
arise when analyzing portfolios of time series, ensembles of climate model outputs, or
collections of spectral measurements where we seek to quantify the shared behavior
across all members rather than merely examining pairs. Moving from pairwise to
multi-way analysis introduces two conceptual choices that shape both the interpretation
and the computational structure of the resulting measures.

The first choice concerns how much of the original signal magnitude we carry into the
coalition analysis. In the multistate setting, each coordinate of a signal is assigned
to exactly one state $B_k$ in a partition $\mathcal{B} = \{B_0,\ldots,B_{K-1}\}$ of
$\mathbb{R}$, with one state (often denoted $B_0$) playing the role of a \emph{neutral}
category. Treating this neutral state on the same footing as the other states leads to
an embedding in which the full $L^1$ norm of each signal is preserved: every unit of
magnitude belongs to exactly one coordinate-state atom. This choice is essential for
coalition accounting, because it allows us to decompose each signal’s total norm into
non-overlapping contributions associated with different coalitions, with no residual
mass left unassigned. In this section we adopt this fully norm-preserving multistate
embedding and use it as the basis for defining cumulative and exclusive intersections
across coalitions.

The second choice involves the granularity of the state partition. A fine multistate
partition (such as distinguishing large losses, small losses, neutral, small gains, and
large gains) provides detailed information about where overlap occurs, but practical
applications often require simpler summaries. We show that a coarse two-state view
(positive versus negative, with neutral optionally reported separately) can be recovered
from fine-grained multistate calculations through principled aggregation, avoiding the
need to perform separate analyses at different resolutions. This dimensional reduction
must be performed carefully: aggregating state channels \emph{before} computing
coordinatewise minima can produce incorrect results, whereas aggregating \emph{after}
computing per-state, per-coordinate minima preserves the intersection structure exactly.

Throughout this section, we consider $m$ signals $A_1, \ldots, A_m \in \mathbb{R}^n$
and analyze how their magnitudes are shared across various subsets, which we term
\emph{coalitions} following the terminology of cooperative game theory. A coalition
$S \subseteq [m]$, where $[m] := \{1, 2, \ldots, m\}$ denotes the index set of all $m$
signals, represents a subset of signals whose collective behavior we wish to quantify.
The singleton coalition $\{j\}$ captures behavior unique to signal $j$, the pair
coalition $\{i,j\}$ captures behavior common to signals $i$ and $j$ but potentially
shared with others, and the grand coalition $[m]$ captures behavior common to all
signals simultaneously. The multistate intersection measures developed below assign a
non-negative magnitude to each coalition, decomposing each signal's total norm into
contributions from the various coalitions to which it belongs.

\subsection{Multistate Embedding with Explicit Neutral State}
\label{ssec:untrimmed_embedding}

We now specialize the multistate, norm-preserving embedding from
Definition~\ref{def:multistate_partition} to the case where one state is treated
as explicitly neutral for coalition analysis. Consider a partition of the real
line into $K$ measurable, pairwise disjoint sets
\begin{equation}
\label{eq:beta_partition}
\mathcal{B} = \{B_0, B_1, \ldots, B_{K-1}\},
\end{equation}
where the indexing begins at zero to emphasize the distinguished role of $B_0$
as a neutral or baseline state. In applications, $B_0$ might represent values
near zero, a reference regime, or any region regarded as qualitatively
different from the remaining states. The sets $B_1, \ldots, B_{K-1}$ then
partition the complement of $B_0$ according to application requirements. For
instance, in financial applications we might define $B_1$ and $B_2$ to represent
large and small losses, while $B_3$ and $B_4$ represent small and large gains.
The key feature of this construction is that all magnitudes, including those in
the neutral state, are retained explicitly rather than being discarded.

Given such a partition $\mathcal{B}$, we simply apply the multistate embedding
$\psi^{(K)}$ of Definition~\ref{def:multistate_partition} to any signal
$A = (A_1,\ldots,A_n) \in \mathbb{R}^n$:
\begin{equation}
[\psi^{(K)}(A)]_{i,k}
\;=\; |A_i| \cdot \mathbf{1}\{A_i \in B_k\},
\qquad i = 1, \ldots, n,\quad k = 0,\ldots,K-1,
\end{equation}
where $\mathbf{1}\{\cdot\}$ denotes the indicator function. No new embedding is
introduced; we simply use $\psi^{(K)}$ with a partition that includes an
explicit neutral state $B_0$.

Because $\{B_k\}_{k=0}^{K-1}$ forms a partition of $\mathbb{R}$, each
coordinate $A_i$ belongs to exactly one state $B_k$ and therefore contributes
its magnitude to exactly one column of the embedded matrix. If $A_i \neq 0$
there is a unique nonzero entry in row $i$, located in the column corresponding
to the state containing $A_i$, with value $|A_i|$; if $A_i = 0$ the entire row
is zero. Summing over states at each coordinate returns $|A_i|$, and summing
these across all coordinates yields the original $L^1$ norm:
\begin{equation}
\label{eq:untrimmed_norm_preservation}
\|A\|_1
= \sum_{i=1}^n |A_i|
= \sum_{i=1}^n \sum_{k=0}^{K-1} [\psi^{(K)}(A)]_{i,k}
= \|\psi^{(K)}(A)\|_1.
\end{equation}
Thus the multistate embedding $\psi^{(K)}$ preserves the full $L^1$ norm of the
signal while recording how each unit of magnitude is allocated across the
qualitative states $B_0,\ldots,B_{K-1}$, including the neutral state.

\subsection{Coalition Structure and Intersection Measures}
\label{ssec:untrimmed_cum_excl}

Having embedded each signal into a multistate representation that preserves
total magnitude, we now define measures that quantify how magnitude is shared
across coalitions of signals. Consider $m$ signals
$A_1, \ldots, A_m \in \mathbb{R}^n$ and their embeddings
$\psi^{(K)}(A_1), \ldots, \psi^{(K)}(A_m)$, each
represented as an $n \times K$ matrix. A coalition $S \subseteq [m]$ is a
nonempty subset of signal indices, representing a group of signals whose
collective behavior we wish to analyze. The power set of $[m]$ contains
$2^m - 1$ nonempty coalitions, ranging from $m$ singleton coalitions
$\{1\}, \{2\}, \ldots, \{m\}$ through $\binom{m}{2}$ pairs and larger groups,
up to the grand coalition $[m]$ containing all signals.

The fundamental quantity we compute for each coalition is the magnitude that
all coalition members share simultaneously while occupying the same state.
This requires examining each state channel separately: at coordinate $i$ and
state $k$, we take the minimum embedded value across all signals in coalition
$S$, representing the largest magnitude that all members simultaneously
contribute to that state--coordinate combination. Summing these minima across
all coordinates and all states yields the total shared magnitude.

\begin{definition}[Cumulative multistate intersection]
\label{def:cumulative_untrimmed}
For signals $A_1, \ldots, A_m \in \mathbb{R}^n$ and a nonempty coalition
$S \subseteq [m]$, the cumulative intersection is defined as
\begin{equation}
\label{eq:U_cumulative}
N(S) := \sum_{k=0}^{K-1} \sum_{i=1}^n
\min_{j \in S} \bigl\{[\psi^{(K)}(A_j)]_{i,k}\bigr\}.
\end{equation}
This quantity aggregates the coordinatewise minimum within each state channel,
summing over all channels to produce a total shared magnitude measured in the
same units as the original signals.
\end{definition}

The structure of this definition merits careful examination. The innermost
operation, $\min_{j \in S}\{[\psi^{(K)}(A_j)]_{i,k}\}$,
identifies the smallest magnitude contributed by any signal in coalition $S$
to the specific combination of coordinate $i$ and state $k$. Since each signal
contributes magnitude to at most one state at each coordinate (and to exactly
one state whenever the underlying coordinate value is nonzero), this minimum
is typically zero except when all coalition members happen to assign
coordinate $i$ to the same state $k$. When such agreement occurs, the minimum
captures the magnitude that all members can simultaneously contribute. The
middle sum over $i = 1, \ldots, n$ aggregates these shared contributions
across all coordinates, and the outer sum over $k = 0, \ldots, K-1$ combines
contributions from all state channels. The resulting scalar $N(S)$
represents the total magnitude that coalition $S$ shares with unanimous state
agreement.

An important property of cumulative intersections is their monotonicity with
respect to coalition inclusion. If $S \subseteq T$, then
$N(T) \leq N(S)$, because taking the minimum over a
larger set of signals cannot increase the result. In particular, the grand
coalition $[m]$ achieves the smallest cumulative intersection, representing
only the magnitude where all $m$ signals simultaneously agree on both
coordinate and state. This monotonicity has an important implication:
cumulative intersections count shared magnitude with multiplicity. Consider a
three-signal example where signals $A_1$ and $A_2$ share substantial magnitude,
and signals $A_2$ and $A_3$ also share substantial magnitude, but the overlap
between all three is small. The pairwise cumulative intersections
$N(\{1,2\})$ and $N(\{2,3\})$ both include the small
three-way overlap in their totals, leading to double-counting if we simply sum
cumulative intersections to allocate budgets.

To obtain non-overlapping contributions that partition each signal's magnitude
exactly, we apply the principle of inclusion--exclusion through Möbius
inversion on the Boolean lattice of coalitions. This mathematical technique,
fundamental in combinatorics and lattice theory, transforms cumulative counts
into exclusive counts by alternately adding and subtracting intersections of
nested coalitions.

\begin{definition}[Exclusive multistate intersection]
\label{def:exclusive_untrimmed}
For a nonempty coalition $S \subseteq [m]$, the exclusive intersection is
obtained by Möbius inversion of the cumulative intersections:
\begin{equation}
\label{eq:U_exclusive}
\widetilde{N}(S) := \sum_{T \supseteq S} \mu(S,T) \, N(T),
\end{equation}
where the Möbius function on the Boolean lattice is given by
$\mu(S,T) = (-1)^{|T| - |S|}$ for $S \subseteq T$ and $\mu(S,T) = 0$ otherwise.
\end{definition}

The Möbius inversion formula assigns to each coalition $S$ the magnitude that is
shared by all members of $S$ but not by any larger coalition. For instance,
$\widetilde{N}(\{1,2\})$ represents magnitude shared by signals
$A_1$ and $A_2$ that is not also shared by any third signal, while
$\widetilde{N}(\{1,2,3\})$ captures magnitude shared by all three
signals simultaneously. The alternating signs in $\mu(S,T) = (-1)^{|T|-|S|}$
implement inclusion--exclusion: we start with the cumulative intersection of
$S$, subtract the cumulative intersections of all coalitions that properly
contain $S$ (correcting for over-counting), add back terms that were subtracted
too many times, and so forth. Standard properties of Möbius inversion on the
Boolean lattice guarantee an exact budget decomposition, and in our min-based,
nonnegative construction the resulting exclusive intersections can be
interpreted as nonnegative ``mass layers'' assigned to each coalition.

\begin{remark}[Computational feasibility of coalition analysis]
We note that computing the exclusive intersection $\overline{N}(S)$ for all $2^m - 1$ nonempty coalitions is computationally exponential in $m$, scaling as $\mathcal{O}(2^m nK)$ for the full decomposition. This exhaustive attribution, while theoretically complete, is often not the practical goal. For many applications, analysis is focused on low-order intersections---such as pairwise $\overline{N}(\{i,j\})$ and three-way $\overline{N}(\{i,j,k\})$ budgets---which remain computationally tractable as $m$ grows. For high-order approximations when $m$ is large, sampling-based methods analogous to Shapley value estimation present a viable path for future investigation.
\end{remark}

\begin{proposition}[Budget identity for exclusive intersections]
\label{prop:budget_identity}
For each signal $A_j$ with $j \in [m]$, the $L^1$ norm decomposes exactly as a
sum of exclusive intersections over all coalitions containing $j$:
\begin{equation}
\|A_j\|_1 = \sum_{S \ni j} \widetilde{N}(S).
\end{equation}
Furthermore, in our min-based construction one can show that each
$\widetilde{N}(S)$ is nonnegative, so the family
$\{\widetilde{N}(S)\}_{S \subseteq [m]}$ forms a bona fide budget
allocation rather than involving artificial cancellations.
\end{proposition}

\begin{remark}[Intuition for non-negativity]
\label{rem:mass_intuition}
Each cumulative intersection $N(S)$ is built from minima of
nonnegative embedded magnitudes. At every coordinate--state pair $(i,k)$, one
can think of each signal as contributing a nonnegative ``stack of mass'', and
$\min_{j \in S}$ picks out the height that all members of $S$ share. Möbius
inversion then peels these shared stacks into layers that are assigned to
different coalitions. Because no negative mass is ever introduced at the
coordinate level, the resulting exclusive intersections
$\widetilde{N}(S)$ can be viewed as nonnegative mass assigned to
coalition $S$. A fully formal derivation is not required for the applications
here; the key point is that the decomposition is both lossless and interpretable.
\end{remark}

\begin{remark}[Partitions as modelling choices, not arbitrary binning]
\label{rem:partition_modelling}
Throughout this work, the multistate partition 
$\mathcal{B} = \{B_0,\ldots,B_{K-1}\}$ is not a generic 
``binning'' device but a \emph{modelling choice} that encodes the 
scientific question of interest. Each state is chosen to represent a 
qualitatively meaningful regime (e.g.\ small vs.\ large gains, mild vs.\ 
extreme temperature, baseline vs.\ elevated insulin use), and the 
peak-to-peak measures then quantify how mass is allocated across these 
regimes.

Consequently, the dependence of $d_{\text{peak}}$ on the partition should 
be interpreted in the same way that dependence on a model class is 
interpreted in classical statistics: different questions call for different 
state resolutions. Coarser partitions answer high-level questions 
(``did we move from low to high volatility?'') while finer partitions 
resolve more detailed hypotheses (``did mass shift specifically from small 
losses to large gains?''). The framework is therefore \emph{hypothesis-driven} 
rather than discretization-driven: changing the partition corresponds to 
changing the question, not to arbitrary numerical binning.

From a technical perspective, working with a state partition is also what 
enables the budget identities. Once a signal is 
represented as nonnegative mass over disjoint coordinate--state atoms 
$(i,k)$, we can apply the Tanimoto/Jaccard construction on the multistate 
embedding to obtain a bounded metric and a positive-semidefinite kernel, and 
perform exact magnitude budgeting across coalitions of signals via Möbius 
inversion. If one were to skip this state layer and work directly with raw 
values, it would no longer be possible to decompose each signal's total 
magnitude $\|A_j\|_1$ into nonnegative coalition budgets that sum exactly 
back to $\|A_j\|_1$.
\end{remark}

\subsection{Illustrative Example: Five-State Partition with Three Signals}
\label{ssec:untrimmed_example}

To make these abstract definitions concrete, consider three financial return signals analyzed with a five-state partition (one neutral and four non-neutral). We use a symmetric partition around zero with a narrow neutral band and a secondary threshold to distinguish moderate from extreme movements. The partition (indexed $k=0,\ldots,4$ so $K=5$ total states) is:
\begin{equation}
\begin{aligned}
B_0 &= [-0.1, 0.1] 
&&\text{(neutral: negligible movement)}, \\
B_1 &= (-\infty, -2.0] 
&&\text{(large loss)}, \\
B_2 &= (-2.0, -0.1) 
&&\text{(small loss)}, \\
B_3 &= (0.1, 2.0] 
&&\text{(small gain)}, \\
B_4 &= (2.0, \infty) 
&&\text{(large gain)}.
\end{aligned}
\end{equation}

This partition creates a symmetric structure around the neutral band, with nested thresholds that separate minor fluctuations from more substantial movements and further distinguish substantial movements from extreme events.

Consider three return signals, each with four temporal observations:
\begin{equation}
A_1 = (8.2, \, 0.3, \, -0.05, \, -5.1), \quad
A_2 = (3.1, \, 0.18, \, -2.6, \, -0.07), \quad
A_3 = (2.2, \, 0.0, \, -1.4, \, -3.0).
\end{equation}
These signals exhibit heterogeneous behavior across coordinates: substantial gains at coordinate $i=1$, modest gains at coordinate $i=2$, values near zero at coordinate $i=3$, and substantial losses at coordinate $i=4$. We now determine the state assignment for each component and construct the embeddings.

\paragraph{State assignments by coordinate.}
At coordinate $i=1$, all three signals report positive returns. Signal $A_1$ has $A_{1,1} = 8.2 > 2.0$, placing it in state $B_4$ (large gain). Signal $A_2$ has $A_{2,1} = 3.1 > 2.0$, also in $B_4$. Signal $A_3$ has $A_{3,1} = 2.2 > 2.0$, again in $B_4$. All three signals therefore contribute to the large-gain state at this coordinate, with respective magnitudes $8.2$, $3.1$, and $2.2$.

At coordinate $i=2$, we observe smaller positive values. Signal $A_1$ has $A_{1,2} = 0.3$, which satisfies $0.1 < 0.3 \leq 2.0$ and therefore belongs to $B_3$ (small gain) with magnitude $0.3$. Signal $A_2$ has $A_{2,2} = 0.18 \in (0.1, 2.0]$, also in $B_3$ with magnitude $0.18$. Signal $A_3$ has $A_{3,2} = 0.0$, which lies in $[-0.1, 0.1]$ and therefore enters $B_0$ (neutral) with magnitude $0.0$. Here two signals agree on small gains while the third signal is neutral.

At coordinate $i=3$, values are negative or near-zero. Signal $A_1$ has $A_{1,3} = -0.05$, which falls in $[-0.1, 0.1]$ and is assigned to $B_0$ (neutral) with magnitude $0.05$. Signal $A_2$ has $A_{2,3} = -2.6$, which satisfies $-2.6 \leq -2.0$ and therefore belongs to $B_1$ (large loss) with magnitude $2.6$. Signal $A_3$ has $A_{3,3} = -1.4 \in (-2.0, -0.1)$, placing it in $B_2$ (small loss) with magnitude $1.4$. All three signals occupy different states at this coordinate, exhibiting no agreement.

At coordinate $i=4$, substantial losses dominate. Signal $A_1$ has $A_{1,4} = -5.1 \leq -2.0$, assigning it to $B_1$ (large loss) with magnitude $5.1$. Signal $A_2$ has $A_{2,4} = -0.07 \in [-0.1, 0.1]$, placing it in $B_0$ (neutral) with magnitude $0.07$. Signal $A_3$ has $A_{3,4} = -3.0 \leq -2.0$, assigning it to $B_1$ (large loss) with magnitude $3.0$. Signals $A_1$ and $A_3$ agree on large losses while $A_2$ remains neutral.

These state assignments determine the structure of the embedded matrices. For signal $A_1$, the embedding $\psi^{(5)}(A_1)$ is a $4 \times 5$ matrix (since we have $n=4$ coordinates and $K=5$ states, with the neutral state arranged in column $0$). Row $1$ has its sole nonzero entry $8.2$ in column $4$ (state $B_4$). Row $2$ has $0.3$ in column $3$ (state $B_3$). Row $3$ has $0.05$ in column $0$ (neutral state $B_0$). Row $4$ has $5.1$ in column $1$ (state $B_1$). The embeddings for $A_2$ and $A_3$ follow analogously from the assignments above.

The $L^1$ norms, computed by summing the absolute values of all coordinates including those in the neutral state, are
\begin{align}
\|A_1\|_1 &= 8.2 + 0.3 + 0.05 + 5.1 = 13.65, \\
\|A_2\|_1 &= 3.1 + 0.18 + 2.6 + 0.07 = 5.95, \\
\|A_3\|_1 &= 2.2 + 0.0 + 1.4 + 3.0 = 6.6.
\end{align}
These norms account for every magnitude in each signal, consistent with the norm preservation property in equation~\eqref{eq:untrimmed_norm_preservation}.

\paragraph{Cumulative intersections for selected coalitions.}
We now compute cumulative intersections for several coalitions to illustrate equation~\eqref{eq:U_cumulative}. Consider first the coalition $S = \{1,2\}$, representing the shared behavior between signals $A_1$ and $A_2$. We evaluate the double sum by examining each state $k$ and each coordinate $i$, computing $\min\{[\psi^{(5)}(A_1)]_{i,k}, [\psi^{(5)}(A_2)]_{i,k}\}$ and summing the results.

At state $B_4$ (large gains), only coordinate $i=1$ has both signals present. The minimum is $\min\{8.2, 3.1\} = 3.1$. At state $B_3$ (small gains), only coordinate $i=2$ has both signals present, yielding $\min\{0.3, 0.18\} = 0.18$. At state $B_1$ (large losses), coordinate $i=4$ has $A_1$ contributing $5.1$ but $A_2$ contributing $0$ (since $A_2$ is in the neutral state at $i=4$), giving minimum $0$. At state $B_2$ (small losses), coordinate $i=3$ has $A_2$ contributing $2.6$ but $A_1$ contributing $0$ (since $A_1$ is neutral at $i=3$), again giving minimum $0$. At the neutral state $B_0$, coordinates $i=3$ and $i=4$ have one or both signals present, but examining the minima: at $i=3$, $\min\{0.05, 0\} = 0$; at $i=4$, $\min\{0, 0.07\} = 0$. Summing all nonzero contributions,
\begin{equation}
N(\{1,2\}) = 3.1 + 0.18 = 3.28.
\end{equation}

Next consider the grand coalition $S = \{1,2,3\}$, requiring agreement among all three signals. At state $B_4$ and coordinate $i=1$, all three signals contribute (magnitudes $8.2$, $3.1$, $2.2$), yielding minimum $\min\{8.2, 3.1, 2.2\} = 2.2$. At state $B_3$ and coordinate $i=2$, signals $A_1$ and $A_2$ contribute ($0.3$ and $0.18$) but $A_3$ contributes $0$ (being in the neutral state), giving minimum $0$. At state $B_1$, coordinate $i=4$ has $A_1$ and $A_3$ contributing ($5.1$ and $3.0$) but $A_2$ contributes $0$, again minimum $0$. At state $B_2$, only $A_3$ contributes at $i=3$, so the minimum is $0$. At the neutral state $B_0$, no coordinate has all three signals simultaneously in the neutral state (at $i=2$, only $A_3$ is neutral; at $i=3$, only $A_1$ is neutral; at $i=4$, only $A_2$ is neutral), producing minimum $0$ at all neutral-state positions. Therefore,
\begin{equation}
N(\{1,2,3\}) = 2.2.
\end{equation}

\paragraph{Exclusive intersections through Möbius inversion.}
The exclusive intersection for coalition $\{1,2\}$ subtracts the cumulative intersection of any coalition properly containing $\{1,2\}$, which is only $\{1,2,3\}$ in this three-signal example. The Möbius function gives $\mu(\{1,2\}, \{1,2\}) = (-1)^{0} = 1$ and $\mu(\{1,2\}, \{1,2,3\}) = (-1)^{1} = -1$, so equation~\eqref{eq:U_exclusive} yields
\begin{equation}
\widetilde{N}(\{1,2\}) 
= N(\{1,2\}) - N(\{1,2,3\}) 
= 3.28 - 2.2 
= 1.08.
\end{equation}
This value represents the magnitude that signals $A_1$ and $A_2$ share exclusively without participation from $A_3$. For the grand coalition, no coalition properly contains $\{1,2,3\}$, so
\begin{equation}
\widetilde{N}(\{1,2,3\}) = N(\{1,2,3\}) = 2.2.
\end{equation}
Similar calculations for other coalitions (singletons, other pairs, and remaining triples if we had more signals) would produce a complete decomposition satisfying Proposition~\ref{prop:budget_identity}. Each signal's norm $\|A_j\|_1$ equals the sum of $\widetilde{N}(S)$ over all coalitions $S$ containing $j$, providing exact budget accounting.

\subsection{Dimensional Reduction: Recovering Two-State Summaries}
\label{ssec:two_state_reduction}

The five-state partition employed in the preceding example provides detailed information about the distribution of gains and losses, distinguishing large from small movements and tracking neutral behavior separately. However, many applications require coarser summaries that aggregate these fine distinctions into broader categories. A common requirement is a simple two-state view separating negative behavior (losses) from positive behavior (gains), potentially with neutral magnitudes reported separately or allocated according to context-specific rules. This section demonstrates how such dimensional reduction can be achieved through post-intersection aggregation of state channels, avoiding the need to recompute intersections from scratch at the coarser resolution.

Define aggregated state groups by partitioning the state indices. Let 
$\mathcal{G}_- \subset \{0,1,\ldots,K-1\}$ denote the collection of loss-state 
indices and $\mathcal{G}_+$ the collection of gain-state indices. In the 
five-state example,
\begin{equation}
\mathcal{G}_- := \{1, 2\} \quad\text{(large loss $B_1$, small loss $B_2$)}, 
\qquad
\mathcal{G}_+ := \{3, 4\} \quad\text{(small gain $B_3$, large gain $B_4$)}.
\end{equation}
The neutral state $B_0$ (index $0$) is handled separately, either reported as a 
third category or omitted from the positive-negative dichotomy depending on 
application needs. The question now is how to compute cumulative intersections for these aggregated categories without re-embedding signals at a coarser resolution.

A naive approach might suggest taking the minimum across signals after first summing over the states within each group. This approach would compute, for coalition $S$, the quantity
\begin{equation}
\text{(INCORRECT)} \quad
\min_{j \in S} \left\{ \sum_{k \in \mathcal{G}_+} \sum_{i=1}^n [\psi^{(K)}(A_j)]_{i,k} \right\}.
\label{eq:wrong_aggregation}
\end{equation}
This formula first aggregates each signal's positive-state magnitudes across all coordinates, producing a single positive-total for each signal, then takes the minimum across the coalition. However, this sequence of operations can severely overestimate the shared magnitude. Consider two signals that both have substantial positive contributions but at different coordinates: signal $A_1$ might have large gains at coordinates $1$ and $2$ but losses at coordinates $3$ and $4$, while signal $A_2$ has losses at $1$ and $2$ but large gains at $3$ and $4$. Summing each signal's positive magnitudes first would produce large totals for both, and taking the minimum would suggest substantial shared positive behavior even though the signals never simultaneously exhibit gains at the same coordinate. This aggregation-then-minimum approach violates the coordinate-wise agreement structure that underlies the intersection concept.

The correct approach maintains coordinate-wise resolution throughout the minimum operation, aggregating state channels only after computing per-coordinate, per-state minima. This ensures that we count only magnitude where coalition members agree on both coordinate and directional category.

\begin{proposition}[Post-intersection state aggregation]
\label{prop:safe_coarsening}
Let $N(S)$ denote the cumulative intersection computed via equation~\eqref{eq:U_cumulative} using the fine partition $\mathcal{B} = \{B_0, B_1, \ldots, B_{K-1}\}$. Define aggregated two-state cumulative intersections by summing the per-coordinate, per-state minima after the minimum operation:
\begin{equation}
N_{+}(S) := \sum_{k \in \mathcal{G}_+} \sum_{i=1}^n \min_{j \in S} \bigl\{[\psi^{(K)}(A_j)]_{i,k}\bigr\}, \quad
N_{-}(S) := \sum_{k \in \mathcal{G}_-} \sum_{i=1}^n \min_{j \in S} \bigl\{[\psi{(K)}(A_j)]_{i,k}\bigr\}.
\end{equation}
Similarly define the neutral contribution as
\begin{equation}
N_{0}(S) := \sum_{i=1}^n \min_{j \in S} \bigl\{[\psi^{(K)}(A_j)]_{i,0}\bigr\}.
\end{equation}
Then $N_{+}(S) + N_{-}(S) + N_{0}(S) = N(S)$, and no intersection mass is lost, gained, or distorted by this aggregation. Furthermore, the aggregated quantities $N_{\pm}(S)$ can be used to define exclusive intersections via Möbius inversion in the usual way, producing a valid two-state budget decomposition.
\end{proposition}

\begin{proof}
By definition, $N(S) = \sum_{k=0}^{K-1} \sum_{i=1}^n \min_{j \in S}\{[\psi^{(K)}(A_j)]_{i,k}\}$. Partitioning the sum over $k$ into the three groups $\{0\}$ (neutral), $\mathcal{G}_-$ (losses), and $\mathcal{G}_+$ (gains) yields $N_{0}(S) + N_{-}(S) + N_{+}(S)$ by definition. Since the min operations are performed before any aggregation, the coordinate-wise agreement structure is preserved exactly, and the sum equals the original fine-grained cumulative intersection. Möbius inversion applied to the aggregated cumulative intersections produces valid exclusive intersections because the Boolean lattice structure depends only on coalition containment, not on the state resolution used to compute intersection magnitudes.
\end{proof}

\begin{remark}[Order of operations is critical]
The proposition emphasizes that state aggregation must occur \emph{after} computing the coordinatewise minima within each state channel. Reversing this order—computing minima after summing states as in equation~\eqref{eq:wrong_aggregation}—produces incorrect results that violate the intersection semantics and can dramatically overestimate shared magnitude. The correct sequence is: (1) compute minima per state and per coordinate, (2) sum over coordinates within each state to get state-specific contributions, (3) aggregate across states as desired. This order respects the fundamental principle that intersection measures quantify simultaneous agreement at each coordinate.
\end{remark}

\paragraph{Two-state summary of the numerical example.}
Applying this aggregation to the computations from Section~\ref{ssec:untrimmed_example}, we extract the positive and negative contributions from coalition $\{1,2\}$. The only nonzero per-state, per-coordinate minima were $3.1$ at state $B_4$ (large gain) and $0.18$ at state $B_3$ (small gain), both of which belong to $\mathcal{G}_+$. Therefore,
\begin{equation}
N_{+}(\{1,2\}) = 3.1 + 0.18 = 3.28, \quad
N_{-}(\{1,2\}) = 0, \quad
N_{0}(\{1,2\}) = 0,
\end{equation}
confirming that $N(\{1,2\}) = 3.28$ arises entirely from shared positive behavior. For the grand coalition $\{1,2,3\}$, the only nonzero contribution was $2.2$ at state $B_4$, yielding
\begin{equation}
N_{+}(\{1,2,3\}) = 2.2, \quad
N_{-}(\{1,2,3\}) = 0, \quad
N_{0}(\{1,2,3\}) = 0.
\end{equation}
The two-state exclusive intersections follow immediately by Möbius inversion:
\begin{equation}
\widetilde{N}_{+}(\{1,2\}) = N_{+}(\{1,2\}) - N_{+}(\{1,2,3\}) = 3.28 - 2.2 = 1.08, \quad
\widetilde{N}_{+}(\{1,2,3\}) = 2.2.
\end{equation}
These values coincide with the full five-state analysis because, in this particular example, all shared magnitude occurred in the positive states. In general, both $N_{+}(S)$ and $N_{-}(S)$ would be nonzero, and each would admit its own Möbius-inverted exclusive decomposition.

\medskip\noindent
\textbf{Summary and interpretation.}
The multistate framework provides complete magnitude accounting through explicit representation of neutral states, cumulative intersections that quantify overlapping behavior across coalitions, and exclusive intersections that partition each signal's magnitude into non-overlapping coalition contributions. When finer state resolution is unnecessary, dimensional reduction through post-intersection aggregation recovers coarser summaries without computational redundancy or loss of mathematical structure. This unified treatment enables analysts to work at the natural resolution for their problem while retaining the flexibility to produce summaries at multiple scales from a single set of calculations.

\subsection{Probabilistic Interpretation via Normalized Embeddings}
\label{sec:prob_interpretation}

The multistate embedding framework developed in the preceding sections admits a natural probabilistic interpretation that connects our geometric similarity measures to classical concepts from probability theory and information theory. This connection provides both theoretical insight—revealing that our distance measures are monotone transformations of total variation distance—and practical utility, as it enables the application of probabilistic reasoning and data-processing inequalities to analyze the behavior of similarity measures under partition refinement or coarsening.

The key observation is that the embedding $\psi^{(K)}(A)$ assigns a non-negative magnitude to each combination of coordinate index $i \in \{1,\ldots,n\}$ and state index $k \in \{0,\ldots,K-1\}$, producing what amounts to a discrete distribution of magnitude across the Cartesian product of coordinates and states. When we normalize by the signal's total $L^1$ norm, this distribution becomes a probability distribution, with the probability mass at each coordinate-state pair proportional to the magnitude that coordinate contributes when occupying that particular state. Comparing two signals then reduces to comparing two probability distributions over the same discrete space, enabling us to invoke the rich theory of probability metrics.

\subsubsection{Construction of the probability space}
\label{ssubsec:prob_interpretation}
We formalize the probabilistic structure by defining an appropriate sample space and measure. Let $\mathcal{B} = \{B_0, \ldots, B_{K-1}\}$ denote a measurable partition of $\mathbb{R}$ as in Section~\ref{ssec:untrimmed_embedding}, and let $\psi^{(K)}$ be the multistate embedding mapping signals $A \in \mathbb{R}^n$ to matrices in $\mathbb{R}_{\geq 0}^{n \times K}$. We construct a finite sample space whose elements represent the possible combinations of coordinate position and state membership.

\begin{definition}[Sample space of coordinate-state pairs]
\label{def:coordinate_state_space}
The sample space of coordinate-state pairs, which we call \emph{atoms} in the sense of atomic probability theory (indivisible elementary events), is defined as the Cartesian product
\begin{equation}
\Omega := \{1, \ldots, n\} \times \{0, \ldots, K-1\}.
\end{equation}
Each element $\omega = (i,k) \in \Omega$ represents the event that we observe coordinate $i$ while it occupies state $k$. The $\sigma$-algebra on $\Omega$ is taken to be the full power set $\mathcal{F} := 2^\Omega$, which is the natural choice for finite spaces and ensures that all subsets are measurable. The space $(\Omega, \mathcal{F})$ thus has exactly $n \cdot K$ atoms, one for each coordinate–state combination.
\end{definition}

The terminology "atom" deserves clarification. In measure theory, an atom is a measurable set of positive measure that cannot be decomposed into smaller measurable sets of positive measure. For finite discrete spaces like $\Omega$, each singleton $\{(i,k)\}$ is an atom, and the entire space is a countable (in this case finite) union of such atoms. This atomic structure means that probability distributions on $(\Omega, \mathcal{F})$ are completely determined by specifying the probability mass assigned to each of the $n \cdot K$ elementary events, with no continuous components or finer internal structure to consider.

For any signal $A \in \mathbb{R}^n$, the multistate embedding $\psi^{(K)}$ naturally induces a finite measure on this space by assigning to each atom the magnitude that $A$ contributes to the corresponding coordinate--state combination.

\begin{definition}[Magnitude measure induced by a signal]
\label{def:magnitude_measure}
For a signal $A \in \mathbb{R}^n$, define the finite measure $\nu_A$ on the measurable space $(\Omega, \mathcal{F})$ by specifying its value on each atom:
\begin{equation}
\nu_A\bigl(\{(i,k)\}\bigr) := [\psi^{(K)}(A)]_{i,k} = |A_i| \cdot \mathbf{1}\{A_i \in B_k\}.
\end{equation}
For arbitrary measurable sets $E \in \mathcal{F}$, extend by countable additivity: $\nu_A(E) = \sum_{\omega \in E} \nu_A(\{\omega\})$. The total measure of the space is
\begin{equation}
\nu_A(\Omega) = \sum_{i=1}^n \sum_{k=0}^{K-1} \nu_A(\{(i,k)\}) = \|A\|_1,
\end{equation}
which follows from equation~\eqref{eq:untrimmed_norm_preservation}.
\end{definition}

The measure $\nu_A$ assigns mass to exactly one atom for each coordinate $i$, 
so at most $n$ atoms carry positive (nonzero) mass. If $A_i = 0$ for some $i$, the 
corresponding atom has zero mass. When the signal $A$ is nonzero, meaning $\|A\|_1 > 0$, we can normalize $\nu_A$ to produce a probability measure.

\begin{definition}[Normalized probability distribution]
\label{def:normalized_probability}
For a nonzero signal $A \in \mathbb{R}^n$ with $\|A\|_1 > 0$, define the probability measure $P_A$ on $(\Omega, \mathcal{F})$ by normalization:
\begin{equation}
P_A := \frac{\nu_A}{\|A\|_1}.
\end{equation}
Explicitly, for each atom $(i,k) \in \Omega$,
\begin{equation}
P_A(\{(i,k)\}) = \frac{[\psi{(K)}(A)]_{i,k}}{\|A\|_1} 
= \frac{|A_i| \cdot \mathbf{1}\{A_i \in B_k\}}{\|A\|_1}.
\end{equation}
This defines a probability distribution over coordinate-state pairs, where the probability of atom $(i,k)$ equals the fraction of signal $A$'s total magnitude contributed by coordinate $i$ when it occupies state $k$.
\end{definition}

For zero signals with $\|A\|_1 = 0$, normalization is undefined, so the 
probabilistic interpretation is restricted to nonzero signals. In practice, 
these degenerate cases are handled directly at the level of the original 
similarity: if both signals are identically zero (or lie entirely within the 
tolerance band), we treat them as perfectly similar, while if exactly one 
signal carries nonzero magnitude, the peak-to-peak similarity reduces to the 
classical Jaccard-style comparison between an empty and a nonempty object. 
These edge cases lie on the boundary of the domain where $\|A\|_1 > 0$ and do 
not affect the probabilistic formulation for nonzero signals.

\subsubsection{Connection to total variation distance}
\label{ssec:tv_connection}

With the probabilistic framework in place, we now make precise the relationship
between the peak-to-peak similarity $J_{\text{peak}}$ (and its associated
distance $d_{\text{peak}}$) and the total variation distance between the
finite measures induced by two signals. The key point is that the original
$J_{\text{peak}}(A,B)$ is defined on \emph{unnormalized} magnitude measures
and therefore depends both on the \emph{shape} and the \emph{total mass}
($L^1$ norm) of the signals. The total variation distance must therefore be
taken between the corresponding finite measures, rather than solely between
the normalized probability distributions.

Recall from Definition~\ref{def:magnitude_measure} that each signal
$A \in \mathbb{R}^n$ induces a finite measure $\nu_A$ on
$(\Omega,\mathcal{F})$ via the multistate embedding $\psi^{(K)}$,
with total mass
\begin{equation}
M_A := \nu_A(\Omega) = \|A\|_1,
\end{equation}
and analogously $M_B := \nu_B(\Omega) = \|B\|_1$ for signal $B$.
The total variation distance between the finite measures $\nu_A$ and $\nu_B$
is defined by
\begin{equation}
\mathrm{TV}(\nu_A,\nu_B)
:= \frac{1}{2} \sum_{\omega \in \Omega}
    \bigl|\nu_A(\{\omega\}) - \nu_B(\{\omega\})\bigr|.
\end{equation}

\begin{proposition}[Peak-to-peak distance as a transform of total variation]
\label{prop:tv_link}
Let $A,B \in \mathbb{R}^n$ and let $\nu_A,\nu_B$ be the finite measures on
$(\Omega,\mathcal{F})$ induced by their multistate embeddings as in
Definition~\ref{def:magnitude_measure}. Denote
\begin{equation}
M_A := \nu_A(\Omega) = \|A\|_1,
\qquad
M_B := \nu_B(\Omega) = \|B\|_1,
\end{equation}

and let
\begin{equation}
\mathrm{TV}(\nu_A,\nu_B)
:= \frac{1}{2} \sum_{\omega \in \Omega}
    \bigl|\nu_A(\{\omega\}) - \nu_B(\{\omega\})\bigr|
\end{equation}
be their total variation distance as finite measures. Then the peak-to-peak
similarity and distance satisfy
\begin{equation}
\label{eq:Jpeak-tv-measure}
J_{\text{peak}}(A,B)
= \frac{M_A + M_B - 2\,\mathrm{TV}(\nu_A,\nu_B)}
       {M_A + M_B + 2\,\mathrm{TV}(\nu_A,\nu_B)},
\end{equation}
and consequently
\begin{equation}
\label{eq:dpeak-tv-measure}
d_{\text{peak}}(A,B)
= 1 - J_{\text{peak}}(A,B)
= \frac{4\,\mathrm{TV}(\nu_A,\nu_B)}
       {M_A + M_B + 2\,\mathrm{TV}(\nu_A,\nu_B)}.
\end{equation}
Equivalently, if we define the normalized discrepancy
\begin{equation}
\label{eq:delta-def}
\delta(A,B)
:= \frac{\mathrm{TV}(\nu_A,\nu_B)}{\tfrac12(M_A + M_B)} \in [0,1],
\end{equation}
then
\begin{equation}
\label{eq:Jpeak-delta}
J_{\text{peak}}(A,B)
= \frac{1 - \delta(A,B)}{1 + \delta(A,B)},
\qquad
d_{\text{peak}}(A,B)
= \frac{2\,\delta(A,B)}{1 + \delta(A,B)}.
\end{equation}
\end{proposition}

\begin{proof}
Write
\begin{equation}
x_\omega := \nu_A(\{\omega\}),\qquad
y_\omega := \nu_B(\{\omega\}),
\end{equation}
so that $x_\omega,y_\omega \ge 0$ for all $\omega \in \Omega$ and
$\sum_{\omega} x_\omega = M_A$, $\sum_{\omega} y_\omega = M_B$.
By Definition~\ref{def:magnitude_measure}, together with the multistate embedding $\psi^{(K)}$, the peak-to-peak similarity is

\begin{equation}
J_{\text{peak}}(A,B)
= \frac{\sum_{\omega \in \Omega} \min\{x_\omega,y_\omega\}}
       {\sum_{\omega \in \Omega} \max\{x_\omega,y_\omega\}}.
\end{equation}

At each atom $\omega \in \Omega$ we have the pointwise identities
\begin{align}
\max\{x_\omega,y_\omega\} + \min\{x_\omega,y_\omega\}
  &= x_\omega + y_\omega, \\
\max\{x_\omega,y_\omega\} - \min\{x_\omega,y_\omega\}
  &= |x_\omega - y_\omega|.
\end{align}
Summing over all $\omega$ gives
\begin{align}
\sum_{\omega} \max\{x_\omega,y_\omega\}
+ \sum_{\omega} \min\{x_\omega,y_\omega\}
&= \sum_{\omega} x_\omega + \sum_{\omega} y_\omega
 = M_A + M_B, \label{eq:sum-plus}\\[0.5em]
\sum_{\omega} \max\{x_\omega,y_\omega\}
- \sum_{\omega} \min\{x_\omega,y_\omega\}
&= \sum_{\omega} |x_\omega - y_\omega|
 = 2\,\mathrm{TV}(\nu_A,\nu_B). \label{eq:sum-minus}
\end{align}
Let
\begin{equation}
S_{\max} := \sum_{\omega} \max\{x_\omega,y_\omega\},
\qquad
S_{\min} := \sum_{\omega} \min\{x_\omega,y_\omega\}.
\end{equation}
Equations~\eqref{eq:sum-plus}–\eqref{eq:sum-minus} form a linear system:
\begin{equation}
S_{\max} + S_{\min} = M_A + M_B,
\qquad
S_{\max} - S_{\min} = 2\,\mathrm{TV}(\nu_A,\nu_B).
\end{equation}
Solving,
\begin{align}
S_{\max}
&= \frac{(M_A + M_B) + 2\,\mathrm{TV}(\nu_A,\nu_B)}{2}, \\
S_{\min}
&= \frac{(M_A + M_B) - 2\,\mathrm{TV}(\nu_A,\nu_B)}{2}.
\end{align}
Substituting into $J_{\text{peak}} = S_{\min}/S_{\max}$ yields
\begin{equation}
J_{\text{peak}}(A,B)
= \frac{M_A + M_B - 2\,\mathrm{TV}(\nu_A,\nu_B)}
       {M_A + M_B + 2\,\mathrm{TV}(\nu_A,\nu_B)},
\end{equation}
which is \eqref{eq:Jpeak-tv-measure}. The distance formula
\eqref{eq:dpeak-tv-measure} follows by computing $1-J_{\text{peak}}$ and
simplifying.

For the normalized discrepancy, note that
\begin{equation}
\delta(A,B)
= \frac{\mathrm{TV}(\nu_A,\nu_B)}{\tfrac12(M_A + M_B)}
= \frac{2\,\mathrm{TV}(\nu_A,\nu_B)}{M_A + M_B},
\end{equation}
so substituting $\mathrm{TV}(\nu_A,\nu_B)
= \tfrac12(M_A + M_B)\,\delta(A,B)$ into \eqref{eq:Jpeak-tv-measure} gives
\begin{equation}
J_{\text{peak}}(A,B)
= \frac{1 - \delta(A,B)}{1 + \delta(A,B)},
\end{equation}
and hence
\begin{equation}
d_{\text{peak}}(A,B)
= 1 - J_{\text{peak}}(A,B)
= \frac{2\,\delta(A,B)}{1 + \delta(A,B)}.
\end{equation}
This completes the proof.
\end{proof}

The representation \eqref{eq:Jpeak-delta} shows that $d_{\text{peak}}$ is a
strictly increasing function of the normalized discrepancy $\delta(A,B)$:
the map $t \mapsto \frac{2t}{1+t}$ has derivative
\begin{equation}
\frac{d}{dt}\left(\frac{2t}{1+t}\right)
= \frac{2}{(1+t)^2} > 0 \quad \text{for } t \in [0,1],
\end{equation}
so larger total variation (relative to average mass) corresponds to larger
peak-to-peak distance.

\begin{remark}[Probability-normalized special case]
\label{rem:prob_tv_link}
When the signals have equal $L^1$ norm, $M_A = M_B > 0$, the finite measures
$\nu_A$ and $\nu_B$ can be written as scaled versions of the probability
measures $P_A$ and $P_B$ of Definition~\ref{def:normalized_probability}:
\begin{equation}
\nu_A = M_A P_A, \qquad \nu_B = M_A P_B.
\end{equation}
In this case,
\begin{equation}
\mathrm{TV}(\nu_A,\nu_B)
= M_A \,\mathrm{TV}(P_A,P_B),
\qquad
\frac{1}{2}(M_A + M_B) = M_A,
\end{equation}
and hence $\delta(A,B) = \mathrm{TV}(P_A,P_B)$. The formulas
\eqref{eq:Jpeak-delta} reduce to
\begin{equation}
J_{\text{peak}}(A,B)
= \frac{1 - \mathrm{TV}(P_A,P_B)}{1 + \mathrm{TV}(P_A,P_B)},
\qquad
d_{\text{peak}}(A,B)
= \frac{2\,\mathrm{TV}(P_A,P_B)}{1 + \mathrm{TV}(P_A,P_B)}.
\end{equation}
Thus, when signals are compared after $L^1$-normalization (or naturally have
equal total magnitude), the peak-to-peak distance becomes a simple monotone
transform of the usual total variation distance between the induced
probability distributions $P_A$ and $P_B$ on the coordinate-state space
$(\Omega,\mathcal{F})$.
\end{remark}

\begin{remark}[Ordinal Transport vs.\ Modular Budgeting]
\label{rem:multistate_cdf}
The multistate partitions $\mathcal{B}=\{B_0,\dots,B_{K-1}\}$ employed in this work (e.g., Eq.~\eqref{eq:beta_partition}) often possess a natural ordinal structure. Consequently, one could define a cumulative magnitude distribution $F_A$ for a signal $A$ by summing the induced measure $\nu_A$ (Def.~4.10) over ordered states:
\[
F_A(k) \;=\; \sum_{j=0}^{k} \nu_A(\{(i, B_j) \mid i \in \{1,\dots,n\}\}), \quad k=0,\dots,K-1.
\]
Constructing distances based on $F_A$—effectively moving from a PDF-like view to a CDF-like view—would shift the geometric framework from Jaccard/Total Variation to the Wasserstein ($W_1$) or Kolmogorov–Smirnov family.

We explicitly avoid this construction to preserve two structural properties essential to the proposed framework. First, the binwise Jaccard geometry adheres to an ``exact-match'' philosophy: the similarity between a \emph{Large Loss} and a \emph{Large Gain} is zero, independent of their ordinal separation. While a transport-based (CDF) metric would penalize this disagreement more heavily than a \emph{Large Loss}/\emph{Small Loss} discrepancy, it introduces gradient sensitivity at the cost of strictly modular accounting. Second, and most critically, the binwise independence of $\psi^{(K)}$ ensures that the space of coalitions forms a Boolean lattice with independent atoms. This independence is the algebraic prerequisite for the exact Möbius budget decomposition derived in Section~\ref{ssec:two_state_reduction}. Cumulative dependencies would couple the states, breaking the lattice structure required for additive coalition budgeting.
\end{remark}

\subsubsection{Data processing inequality and coarsening}
\label{ssec:data_processing}

The probabilistic interpretation in Section~\ref{ssubsec:prob_interpretation} allows us
to view each signal $A$ as inducing a finite measure $\nu_A$ on the
coordinate--state space $\Omega = \{1,\dots,n\}\times\{0,\dots,K-1\}$. On a
finite state space, the total variation distance between two finite measures
$\nu_A$ and $\nu_B$ takes the standard discrete form
\[
\mathrm{TV}(\nu_A,\nu_B)
= \frac12 \sum_{\omega \in \Omega}
   \bigl|\nu_A(\{\omega\}) - \nu_B(\{\omega\})\bigr|,
\]
see, for example, Levin, Peres, and Wilmer~\cite[Ch.~4.1]{LevinPeresWilmer2017}.
This representation makes it transparent that \emph{coarsening} the underlying
partition---by merging atoms of $\Omega$ into larger categories---can only
decrease total variation. The next lemma records the corresponding monotonicity
property for the peak-to-peak distance.

\begin{lemma}[Monotonicity under partition coarsening]
\label{lem:coarsen_monotone}
Let $\pi : \Omega \to \Omega'$ be a measurable mapping that \emph{coarsens} the
atom structure, meaning that $\pi$ maps multiple atoms in $\Omega$ to single
atoms in $\Omega'$, effectively merging coordinate--state categories. For
example, $\pi$ might map both $(i,B_1)$ (large loss) and $(i,B_2)$ (small loss)
to a single atom $(i,\text{loss})$ in a coarser partition.

Let $\nu_A,\nu_B$ be the finite measures induced by signals $A,B$ on
$(\Omega,\mathcal{F})$ as in Definition~\ref{def:magnitude_measure}, and define
their pushforward measures on $(\Omega',\mathcal{F}')$ by
\begin{equation}
\nu_A' := \nu_A \circ \pi^{-1},
\qquad
\nu_B' := \nu_B \circ \pi^{-1}.
\end{equation}
Then
\begin{equation}
\mathrm{TV}(\nu_A', \nu_B') \;\leq\; \mathrm{TV}(\nu_A, \nu_B),
\end{equation}
and consequently, if $d'_{\text{peak}}(A,B)$ denotes the peak-to-peak distance
computed using the coarsened partition,
\begin{equation}
d'_{\text{peak}}(A,B) \;\leq\; d_{\text{peak}}(A,B).
\end{equation}
That is, coarsening the state partition can only decrease the measured distance
(or equivalently, increase the measured similarity).
\end{lemma}

\begin{proof}
On the finite space $\Omega$, total variation admits the discrete
representation
\[
\mathrm{TV}(\nu_A,\nu_B)
= \frac12 \sum_{\omega \in \Omega}
   \bigl|\nu_A(\{\omega\}) - \nu_B(\{\omega\})\bigr|.
\]
Under the coarsening map $\pi$, each coarse atom $\omega' \in \Omega'$ collects
the mass of its preimage $\pi^{-1}(\{\omega'\}) \subseteq \Omega$, so
\[
\nu_A'(\{\omega'\})
= \sum_{\omega \in \pi^{-1}(\{\omega'\})} \nu_A(\{\omega\}),
\qquad
\nu_B'(\{\omega'\})
= \sum_{\omega \in \pi^{-1}(\{\omega'\})} \nu_B(\{\omega\}).
\]
Applying the triangle inequality inside each preimage and summing over
$\omega' \in \Omega'$ shows that merging atoms can only decrease the total
variation, yielding the contraction property
\begin{equation}
\mathrm{TV}(\nu_A', \nu_B') \;\leq\; \mathrm{TV}(\nu_A, \nu_B).
\end{equation}

Let $M_A = \nu_A(\Omega)$ and $M_B = \nu_B(\Omega)$ be the total masses. Since
$\pi$ merely merges atoms, the pushforward measures preserve total mass:
\begin{equation}
\nu_A'(\Omega') = \nu_A(\Omega) = M_A,
\qquad
\nu_B'(\Omega') = \nu_B(\Omega) = M_B.
\end{equation}
Define the normalized discrepancies as in~\eqref{eq:delta-def}:
\begin{equation}
\delta(A,B)
:= \frac{\mathrm{TV}(\nu_A,\nu_B)}{\tfrac12(M_A+M_B)},
\qquad
\delta'(A,B)
:= \frac{\mathrm{TV}(\nu_A',\nu_B')}{\tfrac12(M_A+M_B)}.
\end{equation}
The contraction inequality implies
\begin{equation}
\delta'(A,B) \;\leq\; \delta(A,B).
\end{equation}

By Proposition~\ref{prop:tv_link}, the peak-to-peak distances are given by
\begin{equation}
d_{\text{peak}}(A,B)
= \frac{2\,\delta(A,B)}{1 + \delta(A,B)},
\qquad
d'_{\text{peak}}(A,B)
= \frac{2\,\delta'(A,B)}{1 + \delta'(A,B)},
\end{equation}
cf.~equation~\eqref{eq:Jpeak-delta}. The function
$f(t) = \tfrac{2t}{1+t}$ is strictly increasing on $[0,1]$, since
$f'(t) = \tfrac{2}{(1+t)^2} > 0$, and therefore
\begin{equation}
\delta'(A,B) \leq \delta(A,B)
\quad\Longrightarrow\quad
d'_{\text{peak}}(A,B) \leq d_{\text{peak}}(A,B).
\end{equation}
\end{proof}

\begin{remark}[Refinement increases distance, coarsening decreases distance]
Lemma~\ref{lem:coarsen_monotone} has an important converse implication. While
coarsening partitions decreases distance (and increases similarity), refining
partitions---splitting existing states into finer subcategories---can only
increase the total variation distance and therefore increase $d_{\text{peak}}$
while decreasing $J_{\text{peak}}$. This monotonicity property provides
theoretical justification for the post-intersection aggregation procedure
advocated in Proposition~\ref{prop:safe_coarsening}. When we aggregate state
contributions \emph{after} computing coordinatewise minima, we are effectively
computing the distance at the fine resolution and then reporting aggregated
statistics, which preserves the fine-scale distinction. By contrast, coarsening
\emph{before} computing intersections would compute a fundamentally smaller
distance corresponding to the coarser partition. The choice of partition
granularity thus directly controls the sensitivity of the similarity measure:
finer partitions detect more subtle differences in signal behavior, while
coarser partitions focus on broader patterns and intentionally discard fine
detail.
\end{remark}

\begin{remark}[Zero $L^1$ mass as a physical vacuum]
\label{rem:tv_vacuum}
As in Remark~\ref{rem:zero_state}, the case $\|A\|_1=\|B\|_1=0$
corresponds to the unique ``vacuum'' state with no mass at any coordinate.
In particular, two zero-mass signals represent the same physical regime,
and the peak-to-peak similarity assigns $J_{\text{peak}}(0,0)=1$ and
$d_{\text{peak}}(0,0)=0$. This boundary case lies outside the
normalized-discrepancy formula~\eqref{eq:delta-def}, which assumes
$M_A+M_B>0$, but is fully consistent with the total-variation
representation $TV(\nu_A,\nu_B)=0$ when $\nu_A=\nu_B=0$.
\end{remark}

\subsubsection{Numerical verification of the total variation relationship}

To make the abstract connection between peak-to-peak measures and total variation concrete, we work through a simple numerical example demonstrating that both computational routes---direct min--max calculation versus the total-variation formulation---yield identical results.

\begin{example}[Verification of the TV relationship for $J_{\text{peak}}$]
\label{ex:tv_check}
Consider signals with $n=3$ coordinates and a three-state partition ($K=3$) defined by
\begin{equation}
B_0 = [-0.1, 0.1] \; \text{(neutral)}, \quad
B_1 = (0.1, \infty) \; \text{(positive)}, \quad
B_2 = (-\infty, -0.1) \; \text{(negative)}.
\end{equation}
Let the two signals be
\begin{equation}
A = (2.0, \, 0.3, \, -1.1), \qquad B = (1.6, \, 0.0, \, -0.4).
\end{equation}
We compute the peak-to-peak similarity and distance through two independent routes.

\paragraph{State assignments and norms.}
For signal $A$: coordinate $A_1 = 2.0 > 0.1$ belongs to state $B_1$ with magnitude $2.0$; coordinate $A_2 = 0.3 \in (0.1, \infty)$ belongs to state $B_1$ with magnitude $0.3$; coordinate $A_3 = -1.1 < -0.1$ belongs to state $B_2$ with magnitude $1.1$. Thus
\begin{equation}
\|A\|_1 = 2.0 + 0.3 + 1.1 = 3.4.
\end{equation}

For signal $B$: coordinate $B_1 = 1.6 > 0.1$ belongs to $B_1$ with magnitude $1.6$; coordinate $B_2 = 0.0 \in [-0.1,0.1]$ belongs to $B_0$ with magnitude $0.0$; coordinate $B_3 = -0.4 < -0.1$ belongs to $B_2$ with magnitude $0.4$. Hence
\begin{equation}
\|B\|_1 = 1.6 + 0.0 + 0.4 = 2.0.
\end{equation}

At the level of the magnitude measures $\nu_A, \nu_B$ on $\Omega = \{1,2,3\}\times\{0,1,2\}$, the only atoms with nonzero mass are
\begin{equation}
(1,B_1),\; (2,B_1),\; (3,B_2),
\end{equation}
with
\begin{equation}
\nu_A = (2.0,\, 0.3,\, 1.1), \qquad
\nu_B = (1.6,\, 0.0,\, 0.4)
\end{equation}
in that order.

\paragraph{Route 1: Direct min--max computation.}
At each atom, compute the minimum and maximum magnitudes:

\begin{itemize}
  \item At $(1, B_1)$: $\min\{2.0, 1.6\} = 1.6$, $\max\{2.0, 1.6\} = 2.0$.
  \item At $(2, B_1)$: $\min\{0.3, 0.0\} = 0.0$, $\max\{0.3, 0.0\} = 0.3$.
  \item At $(3, B_2)$: $\min\{1.1, 0.4\} = 0.4$, $\max\{1.1, 0.4\} = 1.1$.
\end{itemize}

Summing across all atoms gives
\begin{align}
\sum_{\omega \in \Omega} \min\{\nu_A(\omega), \nu_B(\omega)\}
&= 1.6 + 0.0 + 0.4 = 2.0, \\
\sum_{\omega \in \Omega} \max\{\nu_A(\omega), \nu_B(\omega)\}
&= 2.0 + 0.3 + 1.1 = 3.4.
\end{align}
Therefore the peak-to-peak similarity and distance are
\begin{equation}
J_{\text{peak}}(A,B) 
= \frac{2.0}{3.4}
= \frac{10}{17} \approx 0.5882, \qquad
d_{\text{peak}}(A,B) 
= 1 - \frac{10}{17}
= \frac{7}{17} \approx 0.4118.
\end{equation}

\paragraph{Route 2: Total-variation formulation.}
Work now at the level of the finite measures $\nu_A,\nu_B$ rather than their individually normalized probability versions. The total-variation distance between $\nu_A$ and $\nu_B$ is
\begin{align}
\mathrm{TV}(\nu_A,\nu_B)
&:= \frac{1}{2} \sum_{\omega \in \Omega} 
\bigl|\nu_A(\omega) - \nu_B(\omega)\bigr| \\
&= \tfrac12\bigl(|2.0-1.6| + |0.3-0.0| + |1.1-0.4|\bigr) \\
&= \tfrac12(0.4 + 0.3 + 0.7)
= \tfrac12 \cdot 1.4
= 0.7.
\end{align}
The average total mass is
\begin{equation}
\frac{\|A\|_1 + \|B\|_1}{2} 
= \frac{3.4 + 2.0}{2}
= 2.7,
\end{equation}
so the normalized discrepancy
\begin{equation}
\delta := \frac{\mathrm{TV}(\nu_A,\nu_B)}{\tfrac12(\|A\|_1 + \|B\|_1)}
= \frac{0.7}{2.7}
= \frac{7}{27}.
\end{equation}
The theoretical relationship between $J_{\text{peak}}$ and this normalized total variation is
\begin{equation}
J_{\text{peak}}(A,B)
= \frac{1 - \delta}{1 + \delta}
= \frac{1 - \frac{7}{27}}{1 + \frac{7}{27}}
= \frac{\tfrac{20}{27}}{\tfrac{34}{27}}
= \frac{20}{34}
= \frac{10}{17},
\end{equation}
which exactly matches the min--max computation above. The corresponding distance is
\begin{equation}
d_{\text{peak}}(A,B)
= 1 - J_{\text{peak}}(A,B)
= \frac{7}{17},
\end{equation}
again agreeing with Route~1.

\medskip\noindent
This example confirms that the peak-to-peak similarity computed directly from magnitudes coincides with the expression obtained via total variation on the induced magnitude measures, provided we use the correct normalized discrepancy
\begin{equation}
\delta = \mathrm{TV}(\nu_A,\nu_B) / \bigl(\tfrac12(\|A\|_1 + \|B\|_1)\bigr)
\end{equation}
rather than total variation between separately normalized probability distributions. Normalizing each signal to a unit-mass probability $P_A, P_B$ would correspond to a different similarity notion, appropriate when we care only about shape and not about total magnitude.
\end{example}

\section{Scalable Coherence Analysis for Large Ensembles}
\label{sec:scalable_coherence}

The Möbius inversion framework of Section~\ref{sec:untrimmed_multistate} provides a
complete $\mathcal{O}(2^m n)$ decomposition of magnitude for all $2^m$ coalitions
of $m$ signals. This is essential for detailed attribution and for understanding
how mass is shared across specific subsets of signals. In contrast to this full
$2^m$-coalition expansion, which is intended for moderate $m$, many applications
simply require a single, scalable $\mathcal{O}(mnK)$ coherence diagnostic:
\emph{``What fraction of the total dynamic range is agreed upon by all $m$
signals simultaneously?''}

This question is answered directly by the Tanimoto similarity of the grand
coalition of all signals. Let $[m] = \{1,\dots,m\}$ denote the index set and
write $S_{\text{all}} = [m]$. Writing $\psi^{(K)}(A_j)$ for the multistate
embedding of signal $A_j$, we define
\begin{equation}
  J_{\text{peak}}(S_{\text{all}})
  \;=\;
  \frac{N(S_{\text{all}})}{U_{\text{peak}}(S_{\text{all}})}
  \;=\;
  \frac{\displaystyle\sum_{\ell=0}^{K-1}\sum_{i=1}^{n}
    \min_{j\in S_{\text{all}}}
    \bigl[\psi^{(K)}(A_{j})\bigr]_{i,\ell}}
       {\displaystyle\sum_{\ell=0}^{K-1}\sum_{i=1}^{n}
    \max_{j\in S_{\text{all}}}
    \bigl[\psi^{(K)}(A_{j})\bigr]_{i,\ell}}.
\end{equation}

The corresponding distance
\begin{equation}
  d_{\text{peak}}(S_{\text{all}}) \;=\; 1 - J_{\text{peak}}(S_{\text{all}})
\end{equation}
quantifies the global disagreement, i.e., the fraction of the ensemble's
dynamic range that is \emph{not} captured by unanimous, sign- and
state-aware agreement across all $m$ signals.

This $\mathcal{O}(mnK)$ grand-coalition measure provides a robust, scalable
diagnostic even for large ensembles (for fixed $K$, the cost is
$\mathcal{O}(mn)$). Furthermore, the per-index components of this
disagreement,
\begin{equation}
  M_i
  \;=\;
  \sum_{\ell=0}^{K-1}
  \left(
    \max_{j\in S_{\text{all}}}
      \bigl[\psi^{(K)}(A_{j})\bigr]_{i,\ell}
    \;-\;
    \min_{j\in S_{\text{all}}}
      \bigl[\psi^{(K)}(A_{j})\bigr]_{i,\ell}
  \right),
  \qquad i = 1,\dots,n,
\end{equation}
can be computed and sorted to pinpoint the coordinates $i$ that contribute
most to the ensemble's incoherence. In this way one obtains a global
coherence score together with an index-wise incoherence profile, all without
incurring the exponential cost of the full $2^m$-coalition Möbius inversion.

\section{Case Study}
\subsection{Sign-Aware Metrics: Phase-Shifted Signal Demonstration}
\label{sec:visualization_case_study}

Consider a synthetic sine wave $A_i = \sin(2\pi t_i/T)$ and a lagged version 
$B_i = \sin(2\pi (t_i-\Delta t)/T)$ sampled at $n=400$ uniformly spaced points 
over one period $T$. For such signals, points near the 
zero-crossings exhibit sign mismatches as one signal leads or lags the other 
through zero.

For phase-shifted sinusoids, the Pearson correlation satisfies 
$r = \cos(\Delta\phi)$, where $\Delta\phi = 2\pi\Delta t/T$ is the phase shift. 
Thus, for a quarter-period shift $\Delta t = T/4$ (a $90^\circ$ phase shift), 
$r = 0$, and for a smaller shift $\Delta t = T/12$ (a $30^\circ$ phase shift), 
$r \approx 0.866$. The sign-aware Jaccard similarity $J_{\text{peak}}$ responds 
differently: with a $90^\circ$ shift we obtain $J_{\text{peak}} \approx 0.17$, 
so the signals retain some overlap where their signs coincide, even though their 
linear correlation is zero. For a $30^\circ$ shift, we obtain 
$J_{\text{peak}} \approx 0.83$, reflecting strong sign-consistent agreement with 
milder penalties from misaligned zero-crossings.

\begin{figure}[tbp]
    \centering
    \includegraphics[width=\textwidth]{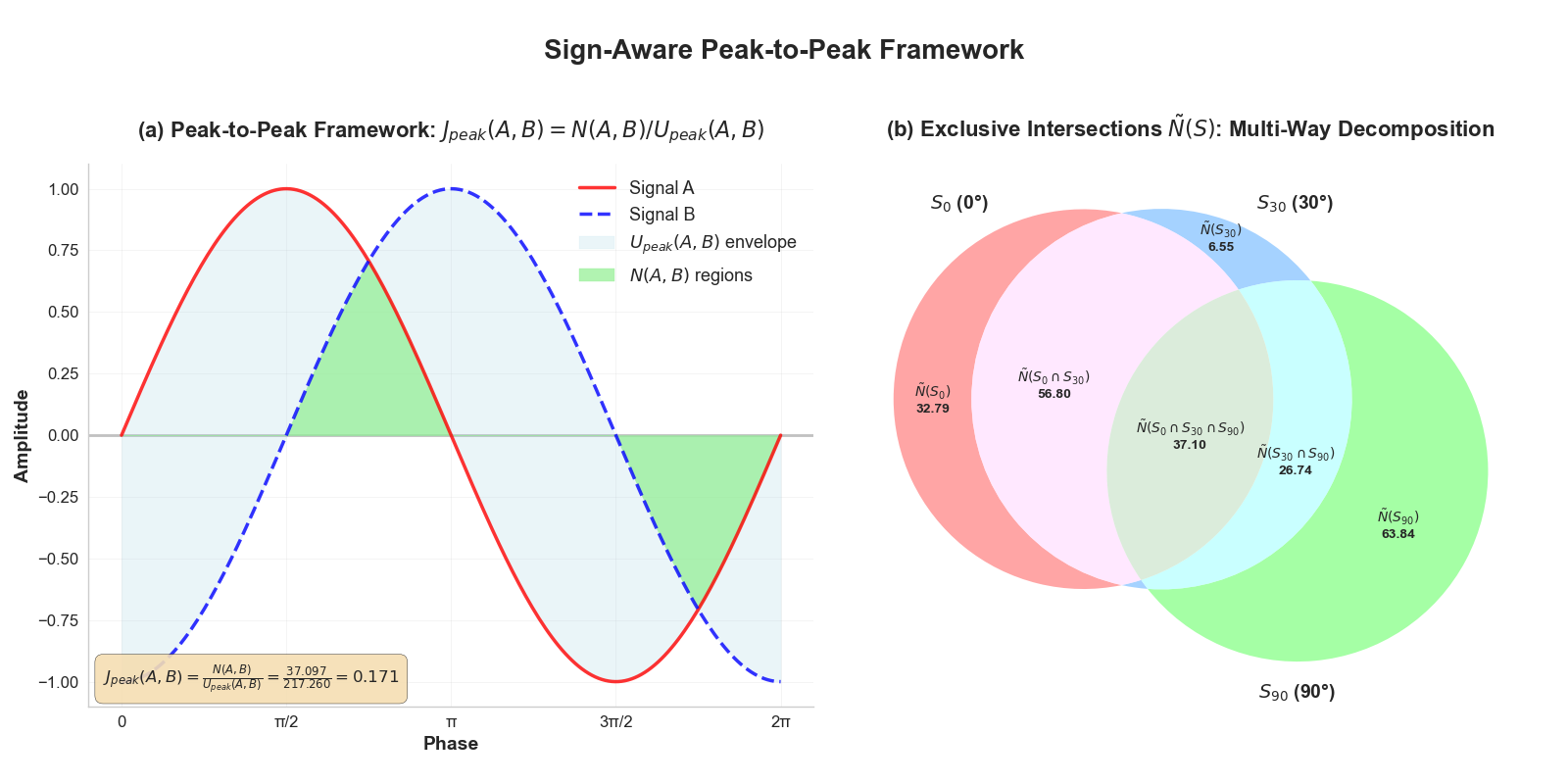} 
    \caption{Visualization of the sign-aware peak-to-peak framework. 
    \textbf{(a)}~Components of the pairwise metric 
    $d_{\text{peak}}(A,B)$ for Signal~A ($\sin(2\pi t/T)$) and 
    Signal~B ($\sin(2\pi(t - \tfrac{T}{4})/T)$) with a $90^\circ$ phase shift. 
    The green shaded area represents the sign-aware intersection $N(A,B)$, and 
    the light blue envelope shows the peak-to-peak union $U_{\text{peak}}(A,B)$ 
    used for normalization. 
    \textbf{(b)}~Venn diagram illustrating the exclusive intersection components 
    $\tilde{N}(S)$ for an ensemble of three signals with phase shifts 
    $0^\circ$, $30^\circ$, and $90^\circ$, demonstrating the multi-way additive 
    decomposition. Numerical labels are in arbitrary amplitude units with peak 
    amplitude normalized to $1$.}
    \label{fig:phase_shift_visualization}
\end{figure}

Figure~\ref{fig:phase_shift_visualization} summarizes the main ideas.  
\textbf{Panel (a)} displays the ingredients of the pairwise distance 
$d_{\text{peak}}(A,B)$ for the $90^\circ$ phase shift: the green band marks the 
sign-consistent intersection $N(A,B)$, while the blue envelope traces the 
peak-to-peak union $U_{\text{peak}}(A,B)$ used for normalization.  
\textbf{Panel (b)} shows the multi-way additive breakdown $\tilde{N}(S)$ for 
three phase-shifted signals, illustrating how shared magnitude is partitioned 
among all coalitions.

Together, the panels demonstrate that the sign-aware framework  
(i)~quantifies agreement strictly where signal polarities concur,  
(ii)~penalizes phase shifts that induce sign reversals---behavior that ordinary 
correlation does not directly register, and  
(iii)~decomposes total magnitude into non-overlapping contributions across 
multiple signals, enabling fine-grained interpretation of multi-signal 
relationships.

\section{Conclusion: Distinctive Properties of the Peak-to-Peak Distance Family}
\label{sec:why_distinctive}

The peak-to-peak distance family developed in this work combines several
well-known mathematical structures: metric distances, positive-semidefinite
kernels, probabilistic interpretations, and exact budget decompositions. None
of these ingredients is new in isolation. What is distinctive is that they all
arise from a single construction based on the sign-split embedding and its
multistate extensions. We briefly summarize the resulting properties and their
joint role in positioning $d_{\text{peak}}$ among existing distance measures.

\subsection{Synthesis of Multiple Mathematical Structures}
\label{ssec:synthesis_property}

The framework is built from three components: (i) the sign-split embedding
$\phi$, which maps signed signals to nonnegative representations while
preserving magnitude; (ii) the multistate embeddings $\psi^{(K)}$ on user
defined partitions of the real line; and (iii) the application of
Tanimoto/Jaccard geometry to these embeddings. From this construction, the
following properties emerge simultaneously.

\paragraph{Bounded, scale-free metric.}
The distance $d_{\text{peak}} : \mathbb{R}^n \times \mathbb{R}^n \to [0,1]$
satisfies all metric axioms (Theorem~\ref{thm:metric_dpeak}). The $[0,1]$ range
makes distances comparable across datasets, and positive homogeneity implies
scale invariance: for all $\alpha>0$,
$d_{\text{peak}}(\alpha A,\alpha B) = d_{\text{peak}}(A,B)$. This bounded,
scale-free metric structure provides a robust foundation for clustering,
hierarchical analysis, and other tasks that rely on well-behaved distances.

\paragraph{Positive-semidefinite kernel.}
The similarity coefficient $J_{\text{peak}}$ defines a positive-semidefinite kernel
$K_{\text{peak}} := J_{\text{peak}}$ (Theorem~\ref{thm:pd_peak_kernel}). For
any finite collection of signals $\{A_1,\dots,A_m\}$, the Gram matrix
$G_{ij} = K_{\text{peak}}(A_i,A_j)$ is positive semidefinite, enabling direct
use in kernel SVMs, kernel PCA, Gaussian processes, and spectral methods
without surrogate feature constructions. The same sign-aware geometry that
defines the distance thus provides a valid kernel for standard ML pipelines.

\paragraph{Probabilistic semantics via normalized embeddings.}
Let $\Omega = \{1,\dots,n\}\times\{0,\dots,K-1\}$ denote the space of
coordinate--state pairs. The multistate embedding $\psi^{(K)}(A)$ induces a
finite measure $\nu_A$ on $\Omega$ with total mass $M_A = \|A\|_1$. Normalizing yields a probability distribution
$P_A = \nu_A/M_A$ whenever $M_A>0$. Proposition~\ref{prop:tv_link} shows that
the discrepancy mass underlying $d_{\text{peak}}$ is exactly
$2\|\nu_A-\nu_B\|_{\mathrm{TV}}$ and that
\begin{equation}
  d_{\text{peak}}(A,B)
  = \frac{2\,\delta(A,B)}{1+\delta(A,B)},
  \qquad
  \delta(A,B)
  := \frac{\mathrm{TV}(\nu_A,\nu_B)}{\tfrac12(M_A+M_B)}.
\end{equation}
In the equal-mass case $M_A=M_B$, this reduces to a monotone transform of
$\mathrm{TV}(P_A,P_B)$. Thus similarity can be read directly in terms of total
variation between induced distributions on $\Omega$, and standard
information-theoretic tools (e.g., data-processing under coarsening) apply.

\paragraph{Exact budget closure via Möbius inversion.}
Because $\nu_A$ is defined on a Boolean algebra of coordinate--state atoms, the
framework admits an exact decomposition of each signal’s magnitude across
coalitions of signals via Möbius inversion
(Definition~\ref{def:exclusive_untrimmed}). For a collection
$\{A_1,\dots,A_m\}$, each $L^1$ norm satisfies
\begin{equation}
  \|A_j\|_1 \;=\; \sum_{S \ni j} \widetilde{N}(S),
\end{equation}
where the sum runs over coalitions containing $j$ and each exclusive
intersection $\widetilde{N}(S)\ge 0$. The decomposition is additive,
unit-preserving, and non-overlapping: every unit of magnitude is assigned to
exactly one coalition with no residual. This budget structure supports
transparent ensemble analysis in settings where magnitude represents physical
or monetary quantities that must be conserved.

\paragraph{Regime–intensity separation.}
The multistate embedding captures both which states are occupied and how much
magnitude each state receives. At the pairwise level, agreement requires both
state alignment (same bin) and comparable magnitudes within that state.
Because $\Omega$ consists of disjoint coordinate--state atoms, the total
discrepancy mass decomposes into two mutually exclusive mechanisms:
cross-state contributions (regime shifts) and within-state contributions
(intensification). This yields an exact global split
\begin{equation}
d_{\text{peak}}(A,B)
\;=\;
\pi_{\mathrm{state}}(A,B) + \pi_{\mathrm{mag}}(A,B),
\end{equation}
where $\pi_{\mathrm{state}}$ measures how much distance is due to state
changes and $\pi_{\mathrm{mag}}$ measures within-state intensity differences.
In applications, this separation allows one to distinguish, for example,
changes in temperature regimes from changes in heatwave intensity, or changes
in spending baskets from changes in spending levels.

\paragraph{Conjunctive distinctiveness.}
Each of the properties above appears in some form in existing frameworks:
cosine similarity and correlation provide kernels but lack metric structure and
budget decompositions; $L^1$ and $L^2$ norms are metrics but ignore sign and
do not yield PSD kernels or multistate budgets; Hellinger distance and
Jensen--Shannon divergence are bounded and probabilistic but require
probability inputs; Wasserstein distances encode geometry but are typically
unbounded and do not provide additive budget partitions. The peak-to-peak
family differs in that a single, sign-aware construction delivers
\emph{simultaneously} a bounded metric, a PSD kernel, probabilistic semantics
linked to total variation, exact coalition budgets, and a built-in
regime–intensity split for signed and complex signals.

\subsection{Comparative Perspective}
\label{ssec:comparative_analysis}

To situate $d_{\text{peak}}$ within the broader landscape, we compared it with
Wasserstein and Hellinger distances along axes such as boundedness, kernel
structure, probabilistic interpretation, and budget behavior
(Table~\ref{tab:distance_comparison}.
These two families represent mature, widely used paradigms in optimal transport
and divergence-based statistics. Relative to them, the peak-to-peak distance
trades some geometric flexibility (e.g., arbitrary ground metrics in
Wasserstein) for a tightly integrated package of properties tailored to signed
and multistate signals. We view this as complementary: $d_{\text{peak}}$
provides a practical default when sign, regime, and magnitude budgeting are
central, while classical distances remain appropriate when other structures
dominate the modeling problem.

{\small
\begin{longtable}{@{}p{3.8cm}p{3.4cm}p{3.4cm}p{3.4cm}@{}}
\caption{Comparative properties of peak-to-peak, Wasserstein, and Hellinger distance families. Each row represents a mathematical property or application requirement, with entries describing whether and how each distance family satisfies that requirement. The peak-to-peak family is distinctive in simultaneously providing all listed properties within a unified construction operating directly on signed real-valued signals.}%
\label{tab:distance_comparison}\\
\toprule
\textbf{Property} & 
\textbf{Peak-to-peak $d_{\text{peak}}$} & 
\textbf{Wasserstein $W_p$} & 
\textbf{Hellinger $H$} \\
\midrule
\endfirsthead

\caption[]{Comparative properties of peak-to-peak, Wasserstein, and Hellinger distance families (continued).}\\
\toprule
\textbf{Property} & 
\textbf{Peak-to-peak $d_{\text{peak}}$} & 
\textbf{Wasserstein $W_p$} & 
\textbf{Hellinger $H$} \\
\midrule
\endhead

\bottomrule
\endfoot

\multicolumn{4}{@{}l}{\textbf{Part I: Core mathematical properties}}\\
\addlinespace[3pt]

\textbf{(a) Bounded metric structure} &
\textbf{Yes}. Distance takes values in $[0,1]$ for all signal pairs, providing scale-free comparisons. Satisfies the triangle inequality. &
\textbf{No} in general. Boundedness requires the underlying ground space to have finite diameter. On $\mathbb{R}^n$ with Euclidean metric, $W_p$ is unbounded. &
\textbf{Yes}. Distance takes values in $[0,1]$ (or $[0,\sqrt{2}]$ depending on normalization). Satisfies the triangle inequality for probability measures. \\
\addlinespace[3pt]

\textbf{(b) Positive semidefinite kernel} &
\textbf{Yes}. The similarity $J_{\text{peak}}$ is a positive-semidefinite kernel on $\mathbb{R}^n$ (Theorem~\ref{thm:pd_peak_kernel}), enabling direct use in SVMs, kernel PCA, and Gaussian processes. &
\textbf{No} in general. While $e^{-\lambda W_p}$ can be positive-semidefinite for specific ground metrics and values of $\lambda$, this is not universal and typically requires case-by-case verification. &
\textbf{No}. Related kernel available. While $H$ itself is a metric rather than a kernel, the Bhattacharyya coefficient $k(p,q) = \sum_i \sqrt{p_i q_i}$ provides a related positive-semidefinite kernel. \\
\addlinespace[3pt]

\textbf{(c) Probabilistic semantics} &
\textbf{Yes}. $d_{\text{peak}}$ is a monotone transform of a normalized total
variation $\delta(A,B)$ between the induced finite measures, with
$d_{\text{peak}} = 2\delta/(1+\delta)$ (Prop. \ref{prop:tv_link}). When $\|A\|_1 = \|B\|_1$,
$\delta$ reduces to $TV(P_A,P_B)$. &
\textbf{Yes}. Wasserstein distances quantify optimal transport cost between
probability measures, with rich geometric and probabilistic interpretation. &
\textbf{Yes}. Hellinger distance is a classical $f$-divergence with
established probabilistic and information-theoretic foundations. \\
\addlinespace[3pt]

\textbf{(d) Exact budget closure} &
\textbf{Yes}. Magnitude decomposes additively across coalitions via Möbius inversion:
$\|A_j\|_1 = \sum_{S \ni j} N_f(S)$ with all terms non-negative (Proposition~\ref{prop:budget_identity}). &
\textbf{No}. Transport cost does not admit additive decomposition into coalition contributions; the distance measures relocation effort rather than partitioning magnitude budgets. &
\textbf{No}. While Hellinger distance can be expressed in terms of probability mass differences, it does not provide an additive budget partition structure. \\
\addlinespace[6pt]

\multicolumn{4}{@{}l}{\textbf{Part II: Practical implementation properties}}\\
\addlinespace[3pt]

\textbf{(e) Mass and intensity control} &
\textbf{Yes}. Multistate channels capture categorical occupancy (mass distribution across states), while magnitude within channels captures quantitative intensity. Safe coarsening preserves structure (Proposition~\ref{prop:safe_coarsening}). &
\textbf{Partial}. Captures mass relocation with geometric intensity via the ground metric (displacement distances), but does not employ explicit multistate channels with independent intensity control per channel. &
\textbf{Partial}. Compares mass distribution across bins in a partition, but lacks a notion of geometric intensity or a built-in safe hierarchical coarsening of state structures. \\
\addlinespace[3pt]

\textbf{(f) Extension to complex signals} &
\textbf{Yes}. Cartesian sign-split embedding treats real and imaginary parts independently. Polar embedding partitions phase angles (Definition~\ref{def:complex_embeddings}). &
\textbf{Yes}. One can define probability measures on $(\mathbb{C}, d)$ for any metric $d$ on the complex plane and compute Wasserstein distances between such measures. &
\textbf{Yes}. Hellinger distance operates on any common measurable space, including $\mathbb{C}$ with an appropriate $\sigma$-algebra. \\
\addlinespace[3pt]

\textbf{(g) Collections and ensembles} &
\textbf{Yes}. Aggregate embeddings over collection members, normalize to $P_{\mathcal{S}}$ on coordinate–state atoms, compare via $d_{\text{peak}}$. Bounded, multiplicity-aware, budget-preserving, and robust to low-mass outliers. &
\textbf{Yes}. Compare empirical or mixture measures induced by the collection. Leverages ground geometry and captures multiplicities, but lacks budget closure. Sensitivity and boundedness depend on the ground metric and domain. &
\textbf{Yes}. Compare mixture distributions on a common $\sigma$-algebra. Bounded and multiplicity-aware, but ignores ground geometric structure. No budget closure or coalition decomposition. \\

\end{longtable}
}

The comparison in Table~\ref{tab:distance_comparison} reveals complementary strengths across the three distance families. Wasserstein distances excel in scenarios where the underlying geometry of the measurement space plays a central role, such as comparing images (where pixel displacement matters) or analyzing distributions on manifolds. The optimal transport framework provides elegant theoretical connections to convex analysis, partial differential equations, and Riemannian geometry. However, this geometric richness comes at the cost of computational complexity (generally requiring the solution of a linear program for discrete measures, or iterative approximation schemes) and the absence, in the generic setting, of certain algebraic structures (bounded range on unbounded domains, positive-semidefinite kernels, budget closure) that are essential for some applications. Hellinger distance, along with other $f$-divergences, provides computationally efficient bounded metrics between probability distributions with strong information-theoretic foundations, but it operates on pre-specified probability measures rather than directly on signal data, and it lacks the budget decomposition structure for coalition analysis.

The peak-to-peak family occupies a distinct niche by providing a complete suite of properties—metric structure, kernel structure, probabilistic semantics, and budget accounting—in a computationally straightforward framework that operates directly on raw signal data (real- or complex-valued, single signals or ensembles) without requiring explicit probability specifications or geometric ground metrics. This makes $d_{\text{peak}}$ particularly well-suited for applications where signals carry physical units (financial returns, atmospheric fluxes, temperature anomalies, spectral coefficients), where both magnitude and directionality matter, where exact accounting is required for interpretability or regulatory compliance, and where kernel-based machine learning methods will be applied to the resulting similarity structure.

\section*{Acknowledgements}

The research was carried out at the Jet Propulsion Laboratory, 
California Institute of Technology, under a contract with the 
National Aeronautics and Space Administration (80NM0018D0004).

\bibliographystyle{apalike} 
\bibliography{references}   

\clearpage
\vfill
\noindent © 2025 California Institute of Technology

\end{document}